\pdfoutput=1
\documentclass{article}
\usepackage[utf8]{inputenc}
\usepackage[T1]{fontenc}
\usepackage{hyperref}
\usepackage{url}
\usepackage{nicefrac}
\usepackage{microtype}
\usepackage{xcolor}
\usepackage{graphicx}
\usepackage{amssymb}
\usepackage{amsmath}
\usepackage{bbold}
\usepackage{algorithm}
\usepackage{algorithmic}
\usepackage[font=small,labelfont=bf,aboveskip=4pt,belowskip=2pt]{caption}
\usepackage{booktabs}
\usepackage{longtable}
\usepackage{multirow}
\usepackage{placeins}
\hypersetup{hidelinks}
\usepackage[main,final]{neurips_2026}

\title{User Embeddings are Superpositions of Interpretable Behavioral Modes}
\author{%
  Jonathan Kozaczuk \quad Zheng Dong \quad Noyan Evirgen \quad Keivan Majidi \quad Eddie K.\ H.\ Ng \\
  \textbf{Chris Porter \quad Qi Zhao \quad Jia Geng \quad Changmao Li \quad Brandon R.\ Feng} \\
  Amazon
}

\begin{document}

\maketitle
\begin{abstract}
User embeddings are central components of modern recommendation, personalization, and behavioral modeling systems. Though these embeddings can compactly encode rich behavioral information, they are inherently dense and opaque, limiting interpretability. To address this gap, we argue that user embeddings can be understood as sparse superpositions of interpretable behavioral modes. This structure is directly predicted by generative models of user behavior for linear embeddings and extends approximately to deep-learning representations by the superposition hypothesis. Superimposed modes can naturally be recovered by sparse dictionary learning methods, such as sparse autoencoders (SAEs), and recovered modes can be automatically named and queried in natural language. We validate mode recovery against known ground truth on synthetic data, where SAEs recover modes better than classical sparse baselines. We then apply our methods to classical and transformer-based user embeddings from two real-world datasets. We also show how the corresponding natural-language labels enable zero-shot audience retrieval and characterization over embeddings never trained on text. The framework is post-hoc and model-agnostic, requiring no modification to the upstream embedding system.

\end{abstract}
\section{Introduction}

Modern personalization, recommendation, and user-modeling systems aim to understand users and tailor experiences to their preferences. Doing so often relies on representing users in the form of \emph{user embeddings}: dense vectors representing their historical actions and preferences that can be compared directly (e.g., for retrieval) or used as inputs to downstream models such as recommenders and LLMs~\citep{raza2025comprehensivereviewrecommendersystems, ning2024userllmefficientllmcontextualization}. These embeddings can be generated in a variety of ways, from classical count matrices to transformer-based sequence encoders~\citep{wang2025embeddingrecommendersystemssurvey}. 

Although user embeddings can encode rich patterns of user behavior, they are not inherently interpretable. This has several important implications. For example, it is not straightforward to identify \emph{why} a specific recommendation is made for a particular user, or audit what information an embedding contains about a user. It is also not typically possible to query the embedding space using natural language as in text-based semantic search. In this work, we show that the superposition framework from mechanistic interpretability (MI) provides a way to understand this opacity and recover transparent behavioral structure.

A key result in MI is that neural network activations can encode human-interpretable concepts in superposition, recoverable by sparse dictionary methods such as sparse autoencoders (SAEs), and auto-interpretable from activating examples~\citep{bricken2023towards, cunningham2024sparse}. We argue and empirically demonstrate that user embeddings encode \emph{behavioral modes} in the same way, making them a natural new test case for the MI framework beyond the transformer-residual-stream settings where it was developed. The resulting sparse decomposition yields an interpretable user representation without per-user labeling or text-based embeddings, enabling applications from user insights to natural language semantic search.

There are other natural approaches for assigning behavioral modes to users. One can define a taxonomy of modes \emph{a priori} and classify each user into a combination, or prompt an LLM to label each user's interaction history directly. Both require raw interaction data for all users, neither operates in the embedding space, and per-user LLM labeling is prohibitively expensive at scale. In contrast, our post-hoc decomposition requires only pre-trained embeddings, discovers modes unsupervised from the data, produces graded mode assignments (e.g., distinguishing a casual from an avid gardener) rather than binary labels, and operates directly in the embedding space.

We validate this framework on synthetic data with known ground-truth modes, where SAEs recover most modes at every degree of superposition, and on two public datasets, where zero-shot natural-language audience retrieval tests the interpretability of the recovered features. Retrieval matches or exceeds classical sparse and text-embedding baselines, supporting the view that real-world user embeddings are sparse superpositions of behavioral modes.

\section{Related Work}\label{sec:related}
\textbf{Sparse autoencoders for embedding interpretability.}
SAEs were first applied to language model activations to extract monosemantic features from polysemantic neurons~\citep{cunningham2024sparse, bricken2023towards}, and have since been scaled to large frontier models~\citep{gao2024scaling, rajamanoharan2024improving}. Recent work extends SAEs to \emph{text} embedding spaces~\citep{ye2024steering, park2025decodingdenseembeddingssparse, kang-etal-2025-interpret, jiang2025interpretableembeddingssparseautoencoders} and to non-LLM embeddings in other domains including speaker, pathology, cell-image, radiology, and CLIP representations~\citep{pluth2025sparse, le2024learning, dasdelen2025cytosae, abdulaal2024xray, zaigrajew2025interpreting}. These methods all ground latents in the semantics available in the input modality (text tokens, images, audio frames). Our work extends SAE decomposition to \emph{behavioral} embeddings that encode non-textual interaction sequences, requiring a different interpretation pipeline and enabling natural language semantic search over embedding spaces that were never trained on text.

\textbf{SAEs for recommender systems.}
Two concurrent works apply SAEs in recommender settings. \citet{spisak2026knots} insert an SAE between the encoder and decoder of a collaborative filtering autoencoder, and \citet{klenitskiy2025sparse} apply SAEs to the hidden states of transformer-based sequential recommenders. Both target recommendation steering or control. Our work differs in that we decompose pre-trained user embeddings post-hoc without modifying the recommendation model, evaluate against known ground-truth behavioral modes, and validate interpretability of the decomposition directly on audience understanding and natural-language semantic search.

\textbf{Criticism of SAEs and random baselines.}
\citet{korznikov2026sanitycheckssparseautoencoders} show that SAEs with frozen-random decoders can match fully trained SAEs in high-dimensional LLM residual streams at high overcompleteness, and \citet{cui2026on} further demonstrates SAEs can fail to recover ground-truth LLM features. We directly address these concerns in Section~\ref{sec:synthetic} by comparing against the soft-frozen-decoder baseline of~\citep{korznikov2026sanitycheckssparseautoencoders} on our synthetic setup, where the ground-truth modes allow a direct measurement of per-user mixture-weight recovery. We find that this equivalence does not hold in our setting at moderate overcompleteness (Section~\ref{sec:synthetic}).

\textbf{User-representation decomposition and disentanglement.}
Prior work on interpreting user representations has taken two approaches. \emph{End-to-end disentanglement}, as in MacridVAE and DisenHAN~\citep{ma2019learning, wang2020disenhan, zheng2021disentangling, fang2025hidvae}, incorporates disentanglement objectives into the upstream recommender at training time. \emph{Post-hoc decomposition} applies unsupervised dictionary learning to pre-trained representations. Classical post-hoc methods include NMF, PCA, Archetypal Analysis, and K-Means for behavior segmentation~\citep{drachen2014comparison}, as well as topic models (LDA and extensions), which model interactions as words drawn from latent topics, each an interpretable distribution over items~\citep{mukherjee2017item, chaudhuri2021modeling}. Our work reflects the post-hoc approach. In contrast to topic models, which operate on raw interactions, our method decomposes pre-trained user embeddings directly, leveraging the richer behavioral signal they encode (e.g., co-viewing sequence, rating intensity, genre co-occurrence) beyond what simple interaction counts capture. Our contribution includes the MI-grounded framework for analyzing user embeddings as sparse superpositions of behavioral concepts, allowing us to recover and identify behavioral modes in natural language. This enables zero-shot natural-language audience retrieval from the recovered modes.

\textbf{Audience retrieval.}
The standard approach to audience retrieval is \emph{lookalike modeling}: given a seed set of example users, find similar users in an embedding space~\citep{liu2019ralm, zhu2021metaheac, peng2023finding, wei2024noah}. These methods require pre-defined example users (``seeds'') and provide no transparency into what behavioral patterns drive the retrieved audience. Our method queries over automatically interpreted sparse behavioral modes, requiring neither seed users nor text-aligned training data.

\section{Superposition from a generative model of user behavior}\label{sec:method-setup}

In this section, we analyze a simple generative model for user behavior to motivate why and how superposition arises in user embeddings. The derivation is exact in expectation for embedding functions that average per-observation features, and serves as a plausibility argument for deep-learning encoders via the superposition hypothesis, which we test empirically for transformer encoders in Sections~\ref{sec:synthetic} and~\ref{sec:real_data} below. Consider a user $\mathcal{U}$ interacting in a specific environment $\mathcal{W}$, where each interaction produces an observation $\mathbf{z} = (z_1, \ldots, z_N)$ comprising $N$ observable features. A given user's behavior in $\mathcal{W}$ is summarized as a matrix $H_\mathcal{U}$, where $(H_\mathcal{U})_{ij} = z_j(\mathcal{U}; t_i)$, the observed feature value for $\mathcal{U}$ at interaction $t_i$. Here $t_i$ labels a particular interaction of $\mathcal{U}$ with the environment. Note that $H_\mathcal{U}$ is implicitly time-dependent, since a user's interactions with the environment are dynamic. We will suppress time dependence throughout this study for simplicity, but at any given snapshot our methodology will apply.

We assume there exist a discrete set of \emph{behavioral modes}, $\{\mathcal{P}_k \mid k = 1, \ldots, K\}$, which are interpretable user types with associated distributions $p_k(\mathbf{z})$ that define the mode's behavior patterns. Modes can be defined at arbitrary granularity: one can always merge two modes into one or split a mode into finer sub-modes. Examples of behavioral modes might be dog lovers on a shopping platform, binge-watchers on a streaming service, or commute-dominated mobility patterns in a ride-sharing app. We further assume each user's behavior is governed by a sparse subset of modes, with user-specific non-negative weights $\mathbf{w} \in \mathbb{R}^K_{\geq 0}$,
$\|\mathbf{w}\|_0 \ll K$, $\sum_k w_k = 1$. Conditioned on $\mathbf{w}$, each interaction for a user $\mathcal{U}$ is drawn independently from the mixture
\begin{equation}\label{eq:mixture}
  p_\mathcal{U}(\mathbf{z}) = \sum_{k=1}^{K} w_k \, p_k(\mathbf{z}).
\end{equation}
The weight vector $\mathbf{w}$ captures both which modes are active and their relative prevalence in the user's behavior. Equation~\eqref{eq:mixture} is a mixed-membership model: each user belongs to multiple modes simultaneously, with each interaction generated by one mode. When the observations are discrete (e.g., individual product purchases treated as tokens), this reduces to the Latent Dirichlet Allocation (LDA) generative model~\citep{blei2003latent}, with behavioral modes playing the role of topics and products playing the role of words. Our formulation generalizes this to continuous and vector-valued observations, but inherits the same core structure, namely a population of users, a shared set of latent modes, and sparse per-user mixtures.

\subsection{Superposition in embedding space}\label{sec:method}

A user embedding $\mathbf{u}$ for user $\mathcal{U}$ is defined by $\mathbf{u} = F(H_\mathcal{U})$, with $F: \mathbb{R}^{T_\mathcal{U} \times N} \rightarrow \mathbb{R}^m$, where $T_\mathcal{U}$ is the user's number of interactions (variable across users), $m$ is the embedding dimension, and typically $m \ll N$. Common examples of $F$ include classical embedding models (e.g., SVD on a counts matrix) or neural networks (e.g., behavioral sequence encoders).

A key consequence of the generative model in Eq.~\eqref{eq:mixture} is that when $F$ is a sample mean of per-observation features $f$ (as with mean-pooled product embeddings), the user embedding inherits a sparse superposition structure in expectation:
\begin{equation}\label{eq:superposition}
  \mathbf{u} \approx \sum_{k=1}^{K} w_k \, \mathbf{d}_k,
  \qquad \text{where}\quad
  \mathbf{d}_k \equiv \mathbb{E}_{z \sim p_k}[f(z)]
\end{equation}
is the expected embedding of observations drawn purely from mode $k$ and $f:\mathbb{R}^N\rightarrow \mathbb{R}^m$ maps individual observations to their embeddings. User embeddings are therefore sparse linear superpositions of \emph{mode centroids} $\mathbf{d}_k$. Analogous expressions hold for linearly-transformed sample-mean pooling, $F = A \cdot \frac{1}{T_\mathcal{U}}\sum_i f(\mathbf{z}_i)$ for a fixed matrix $A$ (e.g., a projection), with the centroids replaced by their images $A\mathbf{d}_k$. These centroids need not be orthogonal: modes with overlapping observation distributions (e.g., ``coffee connoisseur'' and ``home barista'') produce correlated directions, and when $m < K$ even unrelated modes must share directions. Both effects place the centroids in \emph{superposition} in the sense of~\citep{elhage2022superposition}, motivating the use of overcomplete sparse dictionaries rather than orthogonal decompositions for recovery.

Embeddings from \emph{nonlinear} deep-learning encoders admit a similar structure, which we illustrate via the transformer case. Under the superposition hypothesis~\citep{elhage2022superposition}, each residual-stream position satisfies $\mathbf{r}_\ell \approx \sum_j a_j^{(\ell)} \mathbf{d}_j$. For a user embedding formed by any linear pooling of positions, linearity gives $\mathbf{u} \approx \sum_j \tilde{a}_j \mathbf{d}_j$ with $\tilde{a}_j$ the corresponding linear combination of $\{a_j^{(\ell)}\}_\ell$. A mode $\mathcal{P}_k$ induces a distribution over observations and hence over transformer features. If the expected activations decompose linearly in the mode mix $\mathbf{w}$, the feature directions sum into a mode-level centroid $\tilde{\mathbf{d}}_k = \sum_j \mathbb{E}_{p_k}[\tilde{a}_j] \, \mathbf{d}_j$ replacing $\mathbf{d}_k$ in Eq.~\eqref{eq:superposition}, where the expectation is over histories with interactions drawn from $p_k$. Dictionary learning techniques (e.g., sparse autoencoders, discussed below) applied to such pooled embeddings therefore again recover mode-level structure, rather than individual transformer features. For single-position readouts (e.g., last-token, [CLS]) the linear-pooling argument does not apply directly. We conjecture that behavioral-mode resolution emerges from training dynamics, since the objective depends on aggregate sequence structure rather than individual items, but leave verification to future work. Our experiments mean-pool across final-layer positions, so the analysis above applies directly. When attention contextualizes interactions so the linearity-in-$\mathbf{w}$ assumption fails, SAE latents may correspond to finer or composite structure rather than aligning one-to-one with the ground-truth $\mathcal{P}_k$. Since modes admit arbitrary granularity (Section~\ref{sec:method-setup}), the recovered latents remain behavioral modes, at a granularity shaped by the encoder rather than fixed by the generative model.

Under the flat generative model of Eq.~\eqref{eq:mixture} with sample-mean $F$, each $\mathcal{P}_k$ contributes a single direction $\mathbf{d}_k$, so any further splitting of $\mathbf{d}_k$ across multiple SAE latents is artifactual \emph{feature splitting}~\citep{bricken2023towards, templeton2024scaling, chanin2024absorption}. At the single-example level the $L_1$ penalty favors a single aligned atom, but over a population of users the penalty can prefer finer-grained atoms that each activate on a distinct subset of users, reducing the total $L_1$ mass under overcompleteness. Split features can be identified by decoder-direction similarity or correlated activations across users and merged post-hoc, which we validate on synthetic data in Section~\ref{sec:synthetic}. For nonlinear $F$, the previous discussion identified an additional source of finer-grained latents from attention-driven context-dependence; merging absorbs only the artifactual-splitting component.

\subsection{Recovery via sparse autoencoders}\label{sec:recovery}
Equation~\eqref{eq:superposition} implies that recovering behavioral modes from a population of user embeddings is a sparse dictionary learning problem: given $\{\mathbf{u}_i\}_{i=1}^M$, find a dictionary $D \in \mathbb{R}^{m \times d_\text{hidden}}$ and sparse non-negative codes $\{\hat{\mathbf{w}}_i\}$ such that $\mathbf{u}_i \approx D\hat{\mathbf{w}}_i$. The SAE's dictionary width $d_\text{hidden}$ is typically larger than the embedding dimension $m$ to accommodate superposition, and need not equal the ground-truth mode count $K$.

This problem has classical solutions in the discrete observation domain: LDA recovers topics from token counts via variational inference, and NMF recovers non-negative factors from count matrices. However, these methods require access to the raw interaction data and cannot leverage the semantic geometry captured by modern embedding models. We instead operate directly on embeddings using sparse autoencoders (SAEs), which learn a sparse dictionary representation of the input with sparsity enforced by an $L_1$ penalty on encoder activations. In particular, the ``vanilla'' ReLU SAE is defined on an input embedding $\mathbf{u}$ as
\begin{equation}\label{eq:sae}
    \hat{\mathbf{u}} = W_{\rm dec} \, \operatorname{ReLU}(W_{\rm enc} \, \mathbf{u} + \mathbf{b}).
\end{equation}
In terms of Eq.~\eqref{eq:superposition}, the $j$-th SAE latent estimates a mode weight and its decoder column estimates the mode direction:
\begin{align}
&\hat{w}_j \equiv \operatorname{ReLU}(W_{\rm enc} \, \mathbf{u} + \mathbf{b})_j, \qquad
(\hat{\mathbf{d}}_j)_\ell \equiv (W_{\rm dec})_{\ell j}, \quad j = 1, \ldots, d_\text{hidden}.
\end{align}
The SAE is trained to minimize $\mathcal{L}(\mathbf{u}) = \| \hat{\mathbf{u}} - \mathbf{u}\|_2^2 + \lambda \sum_{j=1}^{d_\text{hidden}} \hat{w}_j$, where the sparsity term drops the absolute value since $\hat{w}_j \geq 0$ after the ReLU.
The decoder columns are unit-normalized so that they constitute unit vectors in the original embedding space.

Typically in mechanistic interpretability work, the SAE is \emph{overcomplete}, with the hidden dimension larger than the input dimension ($d_{\rm hidden} > m$), so that superposition can be disentangled. For user embeddings, the number of behavioral modes $K$ may be less than the embedding dimension $m$, however the individual $\mathbf{d}_k$ are still typically in superposition and non-orthogonal. The relevant overcomplete regime in our case is therefore $d_{\rm hidden}$ large enough that the SAE can dedicate features to each mode, or to each sub-mode in the case of nonlinear encoders where behavioral modes split into finer directions. Note that this is not possible with techniques such as PCA, whose components are orthogonal by construction.

SAEs are not the only way to recover sparse modes: in our experiments below we will also compare SAEs against classical sparse dictionary baselines including sparsified versions of PCA, ICA, NMF, and K-Means encoding. Each baseline captures part of the mode structure assumed by Eq.~\eqref{eq:superposition} but lacks at least one key property: overcompleteness (PCA/ICA), the ability to handle signed input (NMF), or the freedom for dictionary directions to point beyond cluster centers (K-Means encoding). We therefore expect SAEs to be the best suited of these methods, although the superposition framework itself does not depend on the choice of decomposition.

Provided the SAE recovers the underlying behavioral modes, its latents should have interpretable descriptions recoverable from the data. We obtain these via auto-interpretation~\citep{bricken2023towards, cunningham2024sparse}: for each SAE latent we pass activating and non-activating user examples with their interaction histories to an LLM and prompt it to identify the dominant pattern in the positives. The output is a set of empirical latents $\{\hat{\mathcal{P}}_k\}$, each comprising a name $\hat{\mathcal{N}}_k$, description $\hat{\mathcal{D}}_k$, and basis vector $\hat{\mathbf{d}}_k$ in the original embedding space. This enables decomposing audiences in terms of behavioral modes, as well as natural-language semantic search by selecting codes with highest semantic overlap with a query and retrieving users with the highest activations on those codes. The same auto-interpretation step can be applied to the classical sparse baselines. Additionally, the LLM outputs a confidence score alongside each name. Because the name and description are grounded in the vocabulary shown to the LLM (e.g., item titles, categories), latents whose defining axis falls outside that vocabulary receive lower confidence. These are typically finer-grained features in the nonlinear-$F$ regime. We expect such latents to remain monosemantic and interpretable under a richer vocabulary, but in practice we gate on confidence ${\geq}0.6$ at retrieval to keep named features within the vocabulary's resolution.

\section{Experiments on Synthetic Data}\label{sec:synthetic}

Assuming behavioral mode superposition in embedding space, to what extent can SAEs and other methods faithfully recover the modes under controlled conditions where the ground truth is known? That the embedding of a sparse mixture is a sparse combination of mode centroids follows from Eq.~\eqref{eq:superposition}. However, it is \emph{not} guaranteed that these centroids can be recovered from the embeddings alone. To test this, we construct synthetic user embeddings with known mode mixtures, apply the SAE decomposition, and measure whether its latents correspond to the true modes (no auto-interpretation is needed, since the ground truth is known). This validates recovery under the framework's assumptions. Whether user embeddings extracted from real-world datasets exhibit the assumed structure is tested in Section~\ref{sec:real_data}. We additionally consider classical baselines (PCA, ICA, K-Means encoding, NMF) to test whether the structural constraints identified in Section~\ref{sec:recovery} (enforced orthogonality, non-negative input, cluster-centered directions) limit their mode recovery in practice.

\paragraph{Setup.}\label{sec:setup}
We define $K{=}75$ behavioral modes (e.g., ``dog lover'', ``new parent''), each with a description and example products. Each of 100{,}000 synthetic users is assigned 2 to 6 active modes, chosen in proportion to mode popularity, with log-normal mixture weights (e.g., 0.7$\times$``dog lover'' + 0.3$\times$``new parent''). We prompt Claude Sonnet 4 to write a purchase history for each user (median 55 orders) whose interactions follow these weights, with 5 to 20\% of the orders drawn as off-mode noise (see Appendix~\ref{app:additionalsynthetic}). We chose $K{=}75$ so that the embedding dimensions swept below ($m = 16$ to $128$) bracket the mode count. The resulting item popularity has the short head and long tail typical of real interaction data, though it is more concentrated than in either real dataset we study (see Appendix~\ref{app:synthetic-stats}). To generate a user embedding from these synthetic interactions, we embed each product with \texttt{all-mpnet-base-v2} and mean-pool over the user's interactions, yielding a 768-dimensional embedding that is a superposition of the underlying modes. To further vary the amount of superposition, we follow the techniques in~\citep{random} and project these embeddings onto random directions in $\mathbb{R}^m$. The dimension $m$ is varied across the experiment. Finally, the embeddings are  standardized and $L_2$-normalized.

To recover the modes, we train an SAE with $d_{\rm hidden} = 300$ ($4{\times}$ overcomplete relative to ground truth) with an adaptive sparsity penalty ensuring that the mean number of active latents ($L_0$) matches the ground-truth mean number of modes per user. We compare against six baselines: random projections onto 300 unit vectors with ReLU, sparsified versions of PCA and FastICA, \emph{K-Means encoding} (cosine similarity to K-Means centroids passed through ReLU and thresholded to match the ground-truth mean $L_0$, following~\citep{coates2012learning}), the soft-frozen decoder SAE of~\citep{korznikov2026sanitycheckssparseautoencoders}, and NMF on shifted (non-negative) embeddings. Matching the ground-truth mode count, K-Means encoding uses 75 components, and PCA, ICA, and NMF use $\min(75, m)$. NMF keeps its natural sparsity (no post-hoc thresholding). K-Means is additionally evaluated with 512 components to isolate raw capacity from decomposition quality. Further details are in Appendix~\ref{app:additionalsynthetic}.

To measure ground truth recovery, each component is matched to its best ground-truth mode by Pearson correlation between component activations and mode weights across users. This mapping is many-to-one: when multiple components map to the same mode, their activations are summed. We also report SAE results after merging features whose decoder directions have cosine similarity ${>}\,0.45$.

\paragraph{Results.}

\begin{figure}[t]
  \centering
  \includegraphics[width=\textwidth]{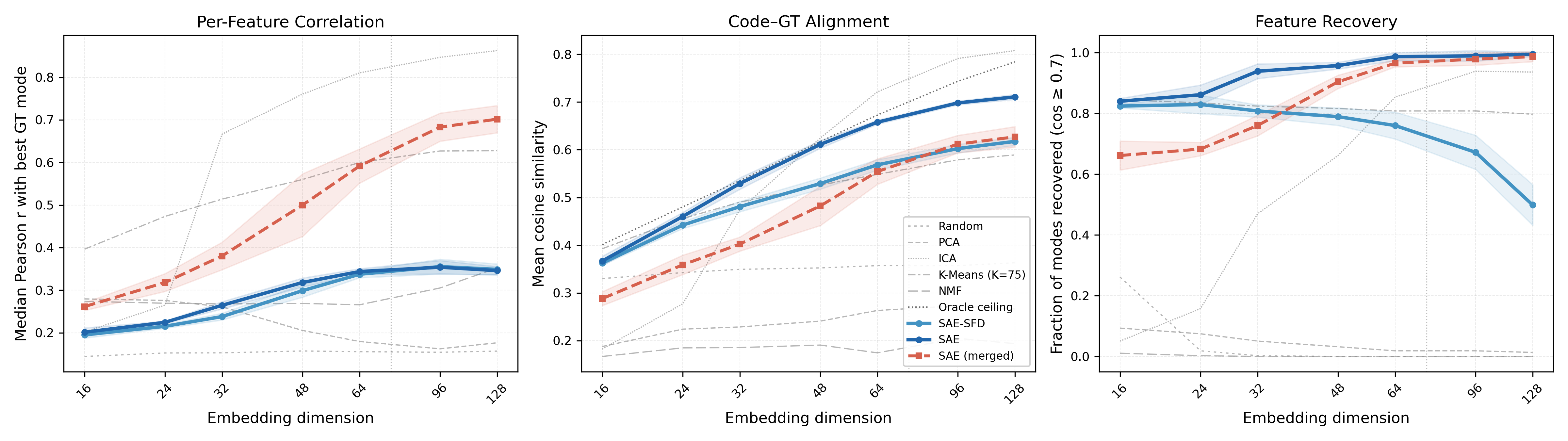}
  \caption{Mode recovery vs.\ embedding dimension $m$ (mean $\pm$ 1 std over 5 random projections). Left: median per-feature Pearson $r$ with best-matching ground-truth mode. Center: mean per-user cosine similarity between projected codes and ground-truth weight vector. Right: fraction of modes recovered at decoder--centroid cosine similarity ${\geq}\,0.7$. The vertical line marks $m = 75$, the ground-truth mode count.}
  \label{fig:comparison_synthetic}
\end{figure}

We evaluate three metrics (Fig.~\ref{fig:comparison_synthetic}, all at matched $L_0$): \emph{per-feature correlation} (Pearson $r$ between each component and its best-matching mode), \emph{code--GT alignment} (per-user cosine similarity after mapping components to modes), and \emph{feature recovery} (fraction of modes with a decoder direction at cosine ${\geq}\,0.7$). An oracle ceiling (least-squares from projected embeddings to ground-truth weights) bounds what any linear method can achieve at each $m$.

SAEs recover most ground-truth modes at every degree of superposition we test, matching or exceeding every classical baseline in Fig.~\ref{fig:comparison_synthetic} on mode recovery. In the high-superposition regime $m < K = 75$, the SAE also stays close to the oracle ceiling on code--GT alignment. K-Means encoding holds steady near $80\%$ recovery (tied with the SAE at $m = 16$) and matches or slightly exceeds the SAE on alignment at $m \leq 24$. PCA recovers almost no modes at any $m$ (${\leq}\,9\%$), as expected since its orthogonal components cannot align with non-orthogonal mode centroids. FastICA overtakes the SAE on alignment from $m = 48$ but recovers fewer modes at cosine ${\geq}\,0.7$ (Fig.~\ref{fig:comparison_synthetic}). K-Means with a 512-component dictionary (not shown in Fig.~\ref{fig:comparison_synthetic}) recovers more mode directions than the SAE at $m \leq 48$ but does not close the alignment gap: it underperforms the 75-component reference at every $m \geq 32$ on code--GT alignment (paired Wilcoxon $p{\approx}0.03$ at $m \geq 48$), so added capacity alone does not give K-Means the SAE's per-user weight recovery. The soft-frozen decoder SAE of~\citep{korznikov2026sanitycheckssparseautoencoders}, their strongest baseline for feature recovery, tracks the SAE only at low $m$ where our setup enters the high-overcompleteness regime ($d_\text{hidden}/m \gg 1$). Note that our real-world data experiments below sit at $d_\text{hidden}/m = 4$ and therefore omit this baseline.

Per-feature correlation reveals the feature splitting discussed in Section~\ref{sec:method}: raw SAE median $r \approx 0.35$ rises to ${\approx}\,0.70$ at $m{=}128$ after merging, confirming the splitting is artifactual. With ground truth known, we can also analyze how monosemantic each feature is: for each mode we identify the feature that best ranks mode-positive users above mode-negative users (Precision among the top $n_\text{pos}$), and report the mean across modes. The merged SAE produces $1.4$--$2\times$ more monosemantic latents than K-Means encoding or raw embedding dimensions ($0.53$ vs.\ $0.39$ and $0.28$ at $m{=}128$, using a separately trained 512-wide SAE on one projection). FastICA attains even higher per-feature monosemanticity ($0.75$) since the generative model's independent-mode assumption exactly matches its objective, but with lower decoder-direction recovery at high cos thresholds (Fig.~\ref{fig:comparison_synthetic}).

\paragraph{Transformer embeddings.} Eq.~\eqref{eq:superposition} holds only approximately for nonlinear encoders (see Section~\ref{sec:method}), so we repeat the previous experiment on embeddings from a trained sequence encoder. We train a transformer sequence recommender (the SASRec encoder of Section~\ref{sec:movielens}) on the synthetic interaction sequences and mean-pool its final-layer hidden states into $d{=}256$ user embeddings (see Appendix~\ref{app:transformer-synthetic}). On the resulting embeddings, the oracle ceiling stays close to that of the averaged product embeddings at the same dimension ($0.78$ vs.\ $0.85$). With the same recovery methods as before, the SAE leads every method on code--GT alignment ($0.58$, vs.\ $0.53$ for FastICA, $0.45$ for K-Means encoding, $0.37$ for LDA fit to the raw user--item counts, and $0.25$ for PCA) and recovers 73 of 75 modes at cosine ${\geq}\,0.7$, close to its $100\%$ on the averaged product embeddings, while FastICA and K-Means encoding drop from $95\%$ to $59\%$ and from $79\%$ to $52\%$. The SAE also recovers more modes than PCA, FastICA, and K-Means encoding at every projected dimension $m \in [16, 128]$, keeps the best alignment when the encoder itself is trained at $d{=}64 < K$, and captures a nearly constant $73$--$75\%$ of the oracle ceiling as the number of modes varies from 15 to 75 (see Appendix~\ref{app:transformer-synthetic}). These experiments suggest that SAE mode recovery carries over from averaged product embeddings to trained transformer embeddings.

\section{Behavioral Modes in Real World Data}\label{sec:real_data}\label{sec:realdata-setup}
We next test whether behavioral mode superposition and recovery carry over to real-world data, where the modes are unknown. We evaluate on two complementary datasets: \texttt{Open e-commerce 1.0}~\citep{berke2024open} (1.8M purchases from 5027 U.S.\ consumers, with a self-reported life-event survey) and MovieLens~1M~\citep{harper2015movielens} (1M ratings, 6040 users, 3706 movies). We use the same pipeline on both: we generate user embeddings (method-specific, described in each subsection), train a sparse decomposition, and auto-interpret each latent with an LLM from its top-activating users. At query time, the feature interpretations are shown to an LLM alongside a natural-language audience query. The LLM scores each latent's relevance on a 0--10 scale and we retain features scoring $\geq 5$. Users are then ranked by the weighted sum of their activations on the retained features. We use Claude Sonnet~4.6 for interpretation and Claude Haiku~4.5 for query-time feature scoring.

Our primary mode recovery method is a ReLU SAE with fixed L1 sparsity penalty and $4\times$ overcompleteness (width chosen per embedding, sparsity $\lambda$ selected via the decoder-orthogonality criterion of~\citep{chanin2025sparse}).\footnote{MovieLens embeddings are $L_2$-normalized and then standardized before SAE training, and Open e-commerce embeddings are used as produced. These steps preserve the form of Eq.~\eqref{eq:superposition} up to a per-user rescaling of the mode weights, a fixed linear map of the centroids, and a constant offset.} As classical sparse alternatives we report K-Means encoding (defined in Section~\ref{sec:synthetic}) and FastICA (the most competitive classical baseline on Open e-commerce), with PCA in Appendix~\ref{app:baselines}. We also report a \emph{text-embedding baseline} as the natural zero-shot NL-retrieval alternative that requires no decomposition: we embed a per-user text summary and the NL query with \texttt{all-MiniLM-L6-v2} and rank by cosine similarity. The summary is a TF-IDF-weighted mean of sentence-transformer embeddings over purchased category names for Open e-commerce (matching the exact signal the SAE, K-Means, and ICA consume) and the mean of watched movie-title embeddings for MovieLens. For each task we report Lift@$n_\text{pos}$, AUROC, Coverage (fraction of positives with nonzero retrieval score), and FVU (fraction of embedding variance unexplained by the best linear decoder fit from the method's codes); Lift, AUROC, and Coverage are averaged across tasks.

\subsection{Open e-commerce}\label{sec:open-ecommerce}

\paragraph{Setup.} First we consider the Open e-commerce dataset. Starting from the raw data, we segment user order histories into 6-month windows, build per-window user-by-category count matrices over the 1864 categories with more than 5 purchases in any window, apply log scaling and TF-IDF weighting, $L_2$-normalize each row, and take truncated SVD to 128 dimensions. Apart from the log scaling and row normalization this embedding is linear in the category counts, so its structure approximately matches Eq.~\eqref{eq:superposition}. The SAE has $d_\text{hidden}{=}512$ and mean $L_0{\approx}23$ (Appendix~\ref{app:ecommerce}). Tasks split into (i)~\emph{category retrieval} (6 tasks spanning base rates 8 to 54\%: pet owner, coffee enthusiast, gardener, board gamer, baby products, musician), where positives are users who purchased in the target categories in post-2021 windows, and (ii)~\emph{life-event retrieval} (new parent, moved residence), where positives are users who self-reported the life event in 2021.

\begin{table}[t]
\caption{Open e-commerce zero-shot NL audience retrieval, metrics averaged per task block (6 category tasks and 2 life-event tasks, see Appendix~\ref{app:ecommerce-tasks}). AUROC$_\text{scored}$ restricts AUROC to users with nonzero scores. Text baseline is the matched-input category-pooled variant (Section~\ref{sec:realdata-setup}). Text baseline has no FVU.}
\label{tab:ecommerce-retrieval}
\centering
\small
\begin{tabular}{llccccc}
\toprule
Task type & Method & FVU $\downarrow$ & Lift & AUROC & AUROC$_\text{scored}$ & Coverage \\
\midrule
\multirow{4}{*}{Category}
  & SAE (ours)                & \textbf{0.075} & \textbf{1.93}$\times$ & \textbf{0.705} & \textbf{0.678} & 0.735 \\
  & FastICA                   & 0.083          & 1.86$\times$          & 0.695          & 0.660          & \textbf{0.770} \\
  & K-Means encoding          & 0.388          & 1.85$\times$          & 0.680          & 0.663          & 0.614 \\
  & Text-embedding baseline   & ---            & 1.39$\times$          & 0.633          & 0.633          & 1.000 \\
\midrule
\multirow{4}{*}{Life-event}
  & SAE (ours)                & \textbf{0.075} & \textbf{5.30}$\times$ & 0.715          & \textbf{0.741} & 0.876 \\
  & FastICA                   & 0.083          & 4.75$\times$          & 0.713          & 0.717          & \textbf{0.932} \\
  & K-Means encoding          & 0.388          & 4.29$\times$          & 0.676          & 0.693          & 0.738 \\
  & Text-embedding baseline   & ---            & 2.13$\times$          & 0.636          & 0.636          & 1.000 \\
\bottomrule
\end{tabular}
\end{table}

\paragraph{Results.} The SAE decomposition recovers the underlying behavioral modes faithfully enough to support zero-shot natural-language retrieval that matches or exceeds every classical sparse baseline on every retrieval-quality metric. It also outperforms the text-embedding baseline on every retrieval-quality metric. The SAE significantly exceeds K-Means encoding on every metric across both task blocks (cross-seed paired Wilcoxon, 5 seeds: all $p{\leq}0.009$; Appendix~\ref{app:stats}) and PCA on every metric (all $p{<}10^{-3}$). FastICA is the closest competitor: on category tasks the SAE has the highest Lift ($p{=}2{\times}10^{-4}$), AUROC ($p{=}0.004$), and AUROC$_\text{scored}$ ($p{=}3{\times}10^{-4}$). On life-event tasks, the SAE has the highest Lift and AUROC$_\text{scored}$ ($p{=}0.001$ each) and ties on AUROC ($p{=}0.5$, $0.715$ vs.\ $0.713$). ICA shows better Coverage across both blocks of tasks (due to its denser activations). The SAE and ICA have the highest reconstruction fidelity (FVU $0.075$ vs.\ $0.083$), while PCA and K-Means lag behind ($0.36$--$0.39$). Note that reconstruction quality does not always translate to retrieval performance: ICA matches SAE on FVU but trails on Lift (Table~\ref{tab:baselines-all}). These results largely hold with an open-weight interpretation model: replacing Claude Sonnet~4.6 with Llama~3.3~70B on one seed leaves SAE retrieval largely unchanged and keeps the SAE ahead of every baseline on mean Lift, with narrower margins over FastICA and K-Means encoding (see Appendix~\ref{app:open-model}).

\subsection{MovieLens 1M}\label{sec:movielens}

\paragraph{Setup.} We next consider MovieLens 1M. We use this dataset to test the \emph{nonlinear} transformer-embedding regime where the superposition of Eq.~\eqref{eq:superposition} holds only approximately (see Section~\ref{sec:method}). To evaluate robustness to encoder architecture we train two transformer recommenders on chronologically ordered interactions: \textbf{SASRec}~\citep{kang2018self} (causal self-attention, next-item prediction) and \textbf{BERT4Rec}~\citep{sun2019bert4rec} (bidirectional self-attention, masked-item prediction), both at $d_\text{model}{=}256$ with user embeddings obtained by mean-pooling final-layer hidden states. The SAE has $d_\text{hidden}{=}1024$ and mean $L_0{\approx}80$. Full SAE and transformer hyperparameters are in Appendix~\ref{app:movielens-sweep}. The decoder-orthogonality operating point (see Section~\ref{sec:real_data}) minimizes artifactual feature splitting in-training, so no post-hoc merging is applied on real data. Residual finer-grained latents under nonlinear $F$ are implicitly handled via the auto-interp confidence filter at retrieval (see Section~\ref{sec:recovery}). Features that track engagement intensity rather than taste are identified and relabeled as engagement-axis features before retrieval, uniformly across SAE and K-Means-encoding codes (Appendix~\ref{app:volume-relabel}). We evaluate on 32 natural-language audience queries (21 broad, 11 niche, full list in Appendix~\ref{app:nl-retrieval}). Ground truth is generated by a deterministic title-pattern rule: for each query we curate a list of matching movie titles from the MovieLens catalog, optionally restricted by MovieLens genre, and label a user positive if their watch history contains $\geq 3$ such titles.

\paragraph{Results.}

\begin{table}[t]
\caption{MovieLens zero-shot NL audience retrieval, averaged over 2 SAE-training seeds $\times$ 32 queries (query list in Appendix~\ref{app:nl-retrieval}). The text baseline is encoder-independent. Bold entries indicate where the SAE significantly outperforms K-Means encoding ($p{<}0.05$, cross-seed paired Wilcoxon; see Appendix~\ref{app:stats}).}
\label{tab:movielens-main}
\centering
\small
\setlength{\tabcolsep}{4pt}
\begin{tabular}{llcccccc}
\toprule
Encoder & Method & FVU $\downarrow$ & Lift & AUROC & AUROC$_\text{scored}$ & Coverage \\
\midrule
\multirow{2}{*}{SASRec}
  & SAE (ours)           & 0.041 & 2.53          & 0.692          & \textbf{0.687} & \textbf{0.977} \\
  & K-Means encoding     & 0.136          & 2.16          & 0.678          & 0.663          & 0.811          \\
\midrule
\multirow{2}{*}{BERT4Rec}
  & SAE (ours)           & 0.032 & 2.38          & 0.690          & 0.688          & \textbf{0.984} \\
  & K-Means encoding     & 0.108          & 2.41          & 0.697          & 0.676          & 0.848          \\
\midrule
--- & Text-embedding baseline & --- & 1.64 & 0.701 & 0.701 & 1.000 \\
\bottomrule
\end{tabular}
\end{table}

The results in Table~\ref{tab:movielens-main} again show that the recovered features support effective zero-shot audience retrieval, consistent with our framework of user embeddings as sparse superpositions of behavioral modes. At matched dictionary size ($d_\text{hidden}{=}1024$) and matched $L_0$, the SAE decomposition and K-Means encoding are matched on Lift and AUROC (cross-seed paired Wilcoxon $p{\geq}0.1$ on both encoders; Table~\ref{tab:stats}). The SAE's advantage is structural, and surfaces on three axes: \emph{coverage} ($98\%$ vs.\ $81$--$85\%$ of true positives scored, $p{<}10^{-4}$ on both encoders), \emph{scored ranking} (higher AUROC$_\text{scored}$ on SASRec, $p{=}0.014$, trending on BERT4Rec, $p{=}0.07$), and \emph{reconstruction fidelity} ($96$--$97\%$ of embedding variance recovered from SAE codes vs.\ $86$--$89\%$ from K-Means codes). The text-embedding baseline reaches $100\%$ coverage by construction (it embeds every user's watched-title history densely) and a comparable AUROC. Its Lift, which measures precision at the top of the ranking, is significantly below the SAE's on both encoders ($p{\leq}0.007$ on each). These results are consistent across both transformer objectives (causal next-item and bidirectional masked). Because MovieLens ground truth is derived from titles, we also retrieve against 14 held-out self-reported demographic labels unseen by any stage of the pipeline. The text-embedding baseline, which matched the SAE on AUROC for the title-derived tasks, falls to near-chance AUROC, significantly below the SAE, and its Lift remains well below the SAE's (see Appendix~\ref{app:demographics}). The coverage advantage holds, and the Lift and AUROC differences with K-Means encoding stay small, at every auto-interpretation confidence threshold from 0.0 to 0.7 (see Appendix~\ref{app:conf-threshold}).

The central MovieLens finding is not that SAE ranks better than K-Means. K-Means encoding is itself a sparse decomposition, and it retrieves competitively because the superposition structure is present for its centroids to approximate. At matched dictionary size and sparsity both methods saturate the same ranking signal, and the finding is \emph{structural}: the SAE dictionary is near-orthogonal (pairwise decoder cosines centered at 0, mean absolute value $0.06$ to $0.08$ on SASRec), while K-Means centroids overlap (mean pairwise cosine $0.29$ on SASRec, $0.17$ on BERT4Rec, see Appendix~\ref{app:movielens-sweep}) and multiple centroids partition the same neighborhood. Overlapping centroids paraphrase one region of the embedding space and leave others uncovered, which costs coverage before it costs rank order, and the same geometry drives the feature-level redundancy in Section~\ref{sec:audience}. On Open e-commerce, whose embedding is approximately linear, the same comparison favors the SAE on every retrieval metric (see Section~\ref{sec:open-ecommerce}). On broad queries against a saturated ranking signal, we do not expect ranking gains from orthogonality alone. Once again, these results suggest that real transformer-encoded embeddings carry superposed behavioral modes that SAEs can recover.

\subsection{Audience characterization}
\label{sec:audience}

A second application of the decomposition is \emph{audience characterization}: given a behaviorally-defined audience, the task is to surface the features that best describe it. A useful top-$k$ feature set should span \emph{distinct} explanatory concepts rather than paraphrase a single concept, and should cover the target audience under semantically-grounded selection. On 20 franchise-level audiences in MovieLens-1M at $k{=}5$, the SAE top-$k$ features are significantly less redundant than the matched-$L_0$ K-Means-encoding top-$k$ ($p{\leq}0.049$ on each encoder) and cover significantly more of the target audience under reverse-selector ranking ($p{\leq}0.0003$ on each encoder, see Appendix~\ref{app:stats}). Within confidence-filtered features ($\geq 0.6$) that surface the same niche audience, K-Means encoding allocates $2.2\times$ more features per distinct concept than SAE ($2.20$ vs $1.00$ redundancy ratio, averaged across 5 niche audiences; Appendix~\ref{app:redundancy}). The effect is even more pronounced on Open e-commerce's approximately linear embeddings ($2.82$ vs $1.14$, or $2.5\times$, see Appendix~\ref{app:redundancy-ecommerce}), as anticipated by our generative framework. The K-Means reference has higher per-feature lift but narrower per-feature activation distributions, so its redundant top-$k$ does not translate into better audience coverage. Figure~\ref{fig:jurassic} illustrates this on the Jurassic Park audience. See Appendix~\ref{app:audience-characterization} for further details related to this analysis.

\begin{figure}[t]
\centering
\footnotesize
\setlength{\tabcolsep}{5pt}
\begin{tabular}{@{}llll@{}}
\toprule
\multicolumn{2}{l}{\textbf{K-Means encoding top-5 for Jurassic Park}} &
\multicolumn{2}{l}{\textbf{SAE (ours) top-5 for Jurassic Park}} \\
\multicolumn{2}{l}{\textit{5/5 paraphrase a comedy concept}} &
\multicolumn{2}{l}{\textit{5 distinct concepts}} \\
\midrule
\textsc{comedy} & Late-1990s indie/quirky dramedies    & \textsc{thriller} & 1990s B-movie action-thriller \\
\textsc{comedy} & Early-2000s indie/quirky comedy      & \textsc{action}   & Low-quality 1990s action blockbuster \\
\textsc{comedy} & Low-budget 1990s B-movie comedy      & \textsc{comedy}   & 1990s low-quality mainstream comedy \\
\textsc{comedy} & 1990s low-quality mainstream comedy  & \textsc{sequel}   & 1980s-90s action-comedy sequel \\
\textsc{comedy} & LGBT romantic comedy and drama       & \textsc{sci-fi}   & 1990s B-grade sci-fi action \\
\bottomrule
\end{tabular}
\caption{Top 5 features describing the Jurassic Park audience (a sci-fi action franchise) under K-Means encoding and SAE on SASRec, ranked by lift at a 10\% top-activator threshold. All five K-Means features paraphrase a mainstream-comedy concept, while the SAE top-5 span five distinct themes. Small-caps tags are our one-word summaries of the LLM-assigned feature names.}
\label{fig:jurassic}
\end{figure}

\section{Conclusion} \label{sec:conclusion}

In this work, we extended the superposition hypothesis for learned representations to user embeddings, arguing that both classical (linear) and deep-learning user embeddings decompose into sparse superpositions of interpretable behavioral modes recoverable by sparse dictionary learning methods such as SAEs. On synthetic data, SAEs recover most of the ground-truth modes even under strong superposition, matching or exceeding classical decomposition methods on both linear embeddings and embeddings learned by a transformer encoder. On the linear embeddings, a fixed post-hoc merging procedure approximately restores the mode granularity assumed by the generative model despite overcomplete feature splitting. On two real-world datasets, where the modes are unknown, the resulting decompositions enable natural-language audience retrieval over embeddings never trained on text, and on MovieLens they describe audiences with more behaviorally distinct features than the strongest classical baseline. Because the method is post-hoc and model-agnostic, it applies to any pre-trained embedding without modifying the upstream system.

More broadly, behavioral mode decomposition provides a general-purpose lens for understanding what user embeddings encode. This has practical implications for auditing recommendation systems, surfacing what behavioral signals drive personalization, and giving practitioners transparent, queryable representations of their user base. The same capability carries privacy risk. Making an embedding queryable in natural language lowers the cost of inferring sensitive traits, such as recent life events, and extends that capability to anyone holding the embeddings and the interaction histories of the example users shown during interpretation. Because the recovered features are enumerable and named, sensitive directions can also be detected, excluded, or ablated, which is not possible with a dense embedding.

Several extensions remain open for future exploration. Our analyses considered fixed embedding snapshots per user, but behavioral modes evolve, and their temporal dynamics, which we did not study, may signal shifts in preference or life circumstance. Our audience retrieval method handles additive concept overlap but not logical constraints or temporal sequencing (e.g., ``viewers who shifted from A to B''). Explicit logical and temporal operations over mode directions are a natural next step. We also experimented only with ReLU SAEs, and other SAE variants and sparse dictionary learning algorithms remain to be tested. Both real-world datasets represent consumption domains with positive-only interaction sequences, and both deep encoders are transformer sequential recommenders (SASRec, BERT4Rec). Domains with different interaction semantics (e.g., mobility or explicit negative feedback) and other deep-learning architectures (e.g., MLP or CNN encoders) remain to be tested, along with single-position readouts (e.g., [CLS] or last-token embeddings). Finally, recent work identifies persona-like structure in LLM internal states via linear probing~\citep{chen2025personavectorsmonitoringcontrolling, lu2026assistantaxissituatingstabilizing}. Our results suggest SAE-based recovery may discover richer persona structure post-hoc, particularly when persona directions are non-orthogonal or overcomplete.

\section*{Acknowledgments}
We thank Yi-Hsuan Kao, Utkrisht Rajkumar, and Siyuan Zhao for insightful discussions.

\bibliographystyle{plainnat}
\bibliography{bib}

\begin{thebibliography}{45}
\providecommand{\natexlab}[1]{#1}
\providecommand{\url}[1]{\texttt{#1}}
\expandafter\ifx\csname urlstyle\endcsname\relax
  \providecommand{\doi}[1]{doi: #1}\else
  \providecommand{\doi}{doi: \begingroup \urlstyle{rm}\Url}\fi

\bibitem[Abdulaal et~al.(2024)Abdulaal, Fry, Monta{\~n}a-Brown, Ijishakin, Gao,
  Hyland, Alexander, and Castro]{abdulaal2024xray}
Ahmed Abdulaal, Hugo Fry, Nina Monta{\~n}a-Brown, Ayodeji Ijishakin, Jack Gao,
  Stephanie Hyland, Daniel~C. Alexander, and Daniel~C. Castro.
\newblock An {X}-ray is worth 15 features: Sparse autoencoders for
  interpretable radiology report generation.
\newblock \emph{arXiv preprint arXiv:2410.03334}, 2024.

\bibitem[Berke et~al.(2024)Berke, Calacci, Mahari, Yabe, Larson, and
  Pentland]{berke2024open}
Alex Berke, Dan Calacci, Robert Mahari, Takahiro Yabe, Kent Larson, and Sandy
  Pentland.
\newblock Open e-commerce 1.0, five years of crowdsourced u.s. amazon purchase
  histories with user demographics.
\newblock \emph{Scientific Data}, 11\penalty0 (1):\penalty0 491, May 2024.
\newblock \doi{10.1038/s41597-024-03329-6}.
\newblock URL \url{https://doi.org/10.1038/s41597-024-03329-6}.

\bibitem[Bingham and Mannila(2001)]{random}
Ella Bingham and Heikki Mannila.
\newblock Random projection in dimensionality reduction: applications to image
  and text data.
\newblock In \emph{Proceedings of the Seventh ACM SIGKDD International
  Conference on Knowledge Discovery and Data Mining}, KDD '01, pages 245--250,
  New York, NY, USA, 2001. Association for Computing Machinery.
\newblock ISBN 158113391X.
\newblock \doi{10.1145/502512.502546}.
\newblock URL \url{https://doi.org/10.1145/502512.502546}.

\bibitem[Blei et~al.(2003)Blei, Ng, and Jordan]{blei2003latent}
David~M. Blei, Andrew~Y. Ng, and Michael~I. Jordan.
\newblock Latent {D}irichlet allocation.
\newblock \emph{Journal of Machine Learning Research}, 3:\penalty0 993--1022,
  2003.

\bibitem[Bricken et~al.(2023)Bricken, Templeton, Batson, Chen, Jermyn, Conerly,
  Turner, Anil, Denison, Askell, Lasenby, Wu, Kravec, Schiefer, Maxwell,
  Joseph, Hatfield-Dodds, Tamkin, Nguyen, McLean, Burke, Hume, Carter,
  Henighan, and Olah]{bricken2023towards}
Trenton Bricken, Adly Templeton, Joshua Batson, Brian Chen, Adam Jermyn, Tom
  Conerly, Nick Turner, Cem Anil, Carson Denison, Amanda Askell, Robert
  Lasenby, Yifan Wu, Shauna Kravec, Nicholas Schiefer, Tim Maxwell, Nicholas
  Joseph, Zac Hatfield-Dodds, Alex Tamkin, Karina Nguyen, Brayden McLean,
  Josiah~E. Burke, Tristan Hume, Shan Carter, Tom Henighan, and Christopher
  Olah.
\newblock Towards monosemanticity: Decomposing language models with dictionary
  learning.
\newblock \emph{Transformer Circuits Thread}, 2023.
\newblock URL
  \url{https://transformer-circuits.pub/2023/monosemantic-features}.

\bibitem[Chanin and Garriga-Alonso(2025)]{chanin2025sparse}
David Chanin and Adri{\`a} Garriga-Alonso.
\newblock {Sparse but Wrong: Incorrect L0 Leads to Incorrect Features in Sparse
  Autoencoders}.
\newblock \emph{arXiv preprint arXiv:2508.16560}, 2025.

\bibitem[Chanin et~al.(2024)Chanin, Wilken-Smith, Dulka, Bhatnagar, and
  Bloom]{chanin2024absorption}
David Chanin, James Wilken-Smith, Tom{\'a}{\v{s}} Dulka, Hardik Bhatnagar, and
  Joseph Bloom.
\newblock A is for absorption: Studying feature splitting and absorption in
  sparse autoencoders.
\newblock \emph{arXiv preprint arXiv:2409.14507}, 2024.

\bibitem[Chaudhuri et~al.(2021)Chaudhuri, Samanta, and
  Sarma]{chaudhuri2021modeling}
Arpita Chaudhuri, Debasis Samanta, and Monalisa Sarma.
\newblock Modeling user behaviour in research paper recommendation system.
\newblock \emph{arXiv preprint arXiv:2107.07831}, 2021.

\bibitem[Chen et~al.(2025)Chen, Arditi, Sleight, Evans, and
  Lindsey]{chen2025personavectorsmonitoringcontrolling}
Runjin Chen, Andy Arditi, Henry Sleight, Owain Evans, and Jack Lindsey.
\newblock Persona vectors: Monitoring and controlling character traits in
  language models.
\newblock \emph{arXiv preprint arXiv:2507.21509}, 2025.

\bibitem[Coates and Ng(2012)]{coates2012learning}
Adam Coates and Andrew~Y. Ng.
\newblock Learning feature representations with {K}-means.
\newblock In Gr{\'e}goire Montavon, Genevieve~B. Orr, and Klaus-Robert
  M{\"u}ller, editors, \emph{Neural Networks: Tricks of the Trade (2nd
  edition)}, volume 7700 of \emph{Lecture Notes in Computer Science}, pages
  561--580. Springer, 2012.

\bibitem[Cui et~al.(2026)Cui, Zhang, Wang, and Wang]{cui2026on}
Jingyi Cui, Qi~Zhang, Yifei Wang, and Yisen Wang.
\newblock On the limits of sparse autoencoders: A theoretical framework and
  reweighted remedy.
\newblock In \emph{The Fourteenth International Conference on Learning
  Representations}, 2026.
\newblock URL \url{https://openreview.net/forum?id=DSOTgzeH3w}.

\bibitem[Cunningham et~al.(2024)Cunningham, Ewart, Riggs, Huben, and
  Sharkey]{cunningham2024sparse}
Hoagy Cunningham, Aidan Ewart, Logan Riggs, Robert Huben, and Lee Sharkey.
\newblock Sparse autoencoders find highly interpretable features in language
  models.
\newblock In \emph{International Conference on Learning Representations
  (ICLR)}, 2024.

\bibitem[Dasdelen et~al.(2025)Dasdelen, Lim, Buck, G{\"o}tze, Marr, and
  Schneider]{dasdelen2025cytosae}
Muhammed~Furkan Dasdelen, Hyesu Lim, Michele Buck, Katharina~S. G{\"o}tze,
  Carsten Marr, and Steffen Schneider.
\newblock {CytoSAE}: Interpretable cell embeddings for hematology.
\newblock \emph{arXiv preprint arXiv:2507.12464}, 2025.

\bibitem[Drachen et~al.(2013)Drachen, Thurau, Sifa, and
  Bauckhage]{drachen2014comparison}
Anders Drachen, Christian Thurau, Rafet Sifa, and Christian Bauckhage.
\newblock A comparison of methods for player clustering via behavioral
  telemetry.
\newblock In \emph{Proceedings of the 8th International Conference on the
  Foundations of Digital Games (FDG)}, pages 245--252, 2013.

\bibitem[Elhage et~al.(2022)Elhage, Hume, Olsson, Schiefer, Henighan, Kravec,
  Hatfield-Dodds, Lasenby, Drain, Chen, Grosse, McCandlish, Kaplan, Amodei,
  Wattenberg, and Olah]{elhage2022superposition}
Nelson Elhage, Tristan Hume, Catherine Olsson, Nicholas Schiefer, Tom Henighan,
  Shauna Kravec, Zac Hatfield-Dodds, Robert Lasenby, Dawn Drain, Carol Chen,
  Roger~B. Grosse, Sam McCandlish, Jared Kaplan, Dario Amodei, Martin
  Wattenberg, and Christopher Olah.
\newblock Toy models of superposition.
\newblock \emph{arXiv preprint arXiv:2209.10652}, 2022.

\bibitem[Fang et~al.(2025)Fang, Gao, Zhu, Li, Zhao, and Chang]{fang2025hidvae}
Dengzhao Fang, Jingtong Gao, Chengcheng Zhu, Yu~Li, Xiangyu Zhao, and Yi~Chang.
\newblock {HiD-VAE}: Interpretable generative recommendation via hierarchical
  and disentangled semantic {IDs}, 2025.
\newblock URL \url{https://arxiv.org/abs/2508.04618}.

\bibitem[Gao et~al.(2024)Gao, Dupr{\'e}~la Tour, Tillman, Goh, Troll, Radford,
  Sutskever, Leike, and Wu]{gao2024scaling}
Leo Gao, Tom Dupr{\'e}~la Tour, Henk Tillman, Gabriel Goh, Rajan Troll, Alec
  Radford, Ilya Sutskever, Jan Leike, and Jeffrey Wu.
\newblock Scaling and evaluating sparse autoencoders.
\newblock \emph{arXiv preprint arXiv:2406.04093}, 2024.

\bibitem[Harper and Konstan(2015)]{harper2015movielens}
F.~Maxwell Harper and Joseph~A. Konstan.
\newblock {The MovieLens Datasets: History and Context}.
\newblock \emph{ACM Transactions on Interactive Intelligent Systems (TiiS)},
  5\penalty0 (4):\penalty0 1--19, 2015.

\bibitem[Jiang et~al.(2025)Jiang, Sun, Dunlap, Smith, and
  Nanda]{jiang2025interpretableembeddingssparseautoencoders}
Nick Jiang, Xiaoqing Sun, Lisa Dunlap, Lewis Smith, and Neel Nanda.
\newblock Interpretable embeddings with sparse autoencoders: A data analysis
  toolkit, 2025.
\newblock URL \url{https://arxiv.org/abs/2512.10092}.

\bibitem[Kang et~al.(2025)Kang, Wang, and Xiong]{kang-etal-2025-interpret}
Hao Kang, Tevin Wang, and Chenyan Xiong.
\newblock Interpret and control dense retrieval with sparse latent features.
\newblock In Luis Chiruzzo, Alan Ritter, and Lu~Wang, editors,
  \emph{Proceedings of the 2025 Conference of the Nations of the Americas
  Chapter of the Association for Computational Linguistics: Human Language
  Technologies (Volume 2: Short Papers)}, pages 700--709, Albuquerque, New
  Mexico, April 2025. Association for Computational Linguistics.
\newblock ISBN 979-8-89176-190-2.
\newblock \doi{10.18653/v1/2025.naacl-short.58}.
\newblock URL \url{https://aclanthology.org/2025.naacl-short.58/}.

\bibitem[Kang and McAuley(2018)]{kang2018self}
Wang-Cheng Kang and Julian McAuley.
\newblock Self-attentive sequential recommendation.
\newblock In \emph{2018 IEEE International Conference on Data Mining (ICDM)},
  pages 197--206. IEEE, 2018.

\bibitem[Klenitskiy et~al.(2025)Klenitskiy, Polev, Denisova, Vasilev, Simakov,
  and Gusev]{klenitskiy2025sparse}
Anton Klenitskiy, Konstantin Polev, Daria Denisova, Alexey Vasilev, Dmitry
  Simakov, and Gleb Gusev.
\newblock Sparse autoencoders for sequential recommendation models:
  Interpretation and flexible control.
\newblock \emph{arXiv preprint arXiv:2507.12202}, 2025.

\bibitem[Korznikov et~al.(2026)Korznikov, Galichin, Dontsov, Rogov, Oseledets,
  and Tutubalina]{korznikov2026sanitycheckssparseautoencoders}
Anton Korznikov, Andrey Galichin, Alexey Dontsov, Oleg Rogov, Ivan Oseledets,
  and Elena Tutubalina.
\newblock {Sanity Checks for Sparse Autoencoders: Do SAEs Beat Random
  Baselines?}, 2026.
\newblock URL \url{https://arxiv.org/abs/2602.14111}.

\bibitem[Le et~al.(2024)Le, Shen, Patel, Shah, Sanghavi, Martin, Eng, Shenker,
  Padigela, Biju, Javed, Hipp, Abel, Pokkalla, Grullon, and
  Juyal]{le2024learning}
Nhat~Minh Le, Ciyue Shen, Neel Patel, Chintan Shah, Darpan Sanghavi, Blake
  Martin, Alfred Eng, Daniel Shenker, Harshith Padigela, Raymond Biju,
  Syed~Ashar Javed, Jennifer Hipp, John Abel, Harsha Pokkalla, Sean Grullon,
  and Dinkar Juyal.
\newblock Learning biologically relevant features in a pathology foundation
  model using sparse autoencoders.
\newblock \emph{arXiv preprint arXiv:2407.10785}, 2024.

\bibitem[Liu et~al.(2019)Liu, Ge, Zhang, and Lin]{liu2019ralm}
Yudan Liu, Kaikai Ge, Xu~Zhang, and Leyu Lin.
\newblock Real-time attention based look-alike model for recommender system.
\newblock In \emph{Proceedings of the 25th ACM SIGKDD International Conference
  on Knowledge Discovery \& Data Mining}, 2019.

\bibitem[Lu et~al.(2026)Lu, Gallagher, Michala, Fish, and
  Lindsey]{lu2026assistantaxissituatingstabilizing}
Christina Lu, Jack Gallagher, Jonathan Michala, Kyle Fish, and Jack Lindsey.
\newblock The assistant axis: Situating and stabilizing the default persona of
  language models.
\newblock \emph{arXiv preprint arXiv:2601.10387}, 2026.

\bibitem[Ma et~al.(2019)Ma, Zhou, Cui, Yang, and Zhu]{ma2019learning}
Jianxin Ma, Chang Zhou, Peng Cui, Hongxia Yang, and Wenwu Zhu.
\newblock Learning disentangled representations for recommendation.
\newblock In \emph{Advances in Neural Information Processing Systems
  (NeurIPS)}, volume~32, 2019.

\bibitem[Mukherjee et~al.(2017)Mukherjee, Lamba, and Weikum]{mukherjee2017item}
Subhabrata Mukherjee, Hemank Lamba, and Gerhard Weikum.
\newblock Item recommendation with evolving user preferences and experience.
\newblock \emph{arXiv preprint arXiv:1705.02519}, 2017.

\bibitem[Ning et~al.(2024)Ning, Liu, Wu, Wu, Berlowitz, Prakash, Green,
  O'Banion, and Xie]{ning2024userllmefficientllmcontextualization}
Lin Ning, Luyang Liu, Jiaxing Wu, Neo Wu, Devora Berlowitz, Sushant Prakash,
  Bradley Green, Shawn O'Banion, and Jun Xie.
\newblock {User-LLM: Efficient LLM Contextualization with User Embeddings},
  2024.
\newblock URL \url{https://arxiv.org/abs/2402.13598}.

\bibitem[Park et~al.(2025)Park, Kim, and
  Ko]{park2025decodingdenseembeddingssparse}
Seongwan Park, Taeklim Kim, and Youngjoong Ko.
\newblock Decoding dense embeddings: Sparse autoencoders for interpreting and
  discretizing dense retrieval, 2025.
\newblock URL \url{https://arxiv.org/abs/2506.00041}.

\bibitem[Peng et~al.(2023)Peng, Liu, and Shen]{peng2023finding}
Yang Peng, Changzheng Liu, and Wei Shen.
\newblock Finding lookalike customers for e-commerce marketing.
\newblock \emph{arXiv preprint arXiv:2301.03147}, 2023.

\bibitem[Pluth et~al.(2025)Pluth, Zhou, and Gurbani]{pluth2025sparse}
Daniel Pluth, Yu~Zhou, and Vijay~K. Gurbani.
\newblock Sparse autoencoder insights on voice embeddings.
\newblock \emph{arXiv preprint arXiv:2502.00127}, 2025.

\bibitem[Rajamanoharan et~al.(2024)Rajamanoharan, Conmy, Smith, Lieberum,
  Varma, Kram{\'a}r, Shah, and Nanda]{rajamanoharan2024improving}
Senthooran Rajamanoharan, Arthur Conmy, Lewis Smith, Tom Lieberum, Vikrant
  Varma, J{\'a}nos Kram{\'a}r, Rohin Shah, and Neel Nanda.
\newblock Improving dictionary learning with gated sparse autoencoders.
\newblock \emph{arXiv preprint arXiv:2404.16014}, 2024.

\bibitem[Raza et~al.(2025)Raza, Rahman, Kamawal, Toroghi, Raval, Navah, and
  Kazemeini]{raza2025comprehensivereviewrecommendersystems}
Shaina Raza, Mizanur Rahman, Safiullah Kamawal, Armin Toroghi, Ananya Raval,
  Farshad Navah, and Amirmohammad Kazemeini.
\newblock A comprehensive review of recommender systems: Transitioning from
  theory to practice, 2025.
\newblock URL \url{https://arxiv.org/abs/2407.13699}.

\bibitem[Sp{\'\i}{\v{s}}{\'a}k et~al.(2026)Sp{\'\i}{\v{s}}{\'a}k, Pe{\v{s}}ka,
  {\v{S}}koda, Van{\v{c}}ura, and Alves]{spisak2026knots}
Martin Sp{\'\i}{\v{s}}{\'a}k, Ladislav Pe{\v{s}}ka, Petr {\v{S}}koda,
  Vojt{\v{e}}ch Van{\v{c}}ura, and Rodrigo Alves.
\newblock From knots to knobs: Towards steerable collaborative filtering using
  sparse autoencoders.
\newblock \emph{arXiv preprint arXiv:2601.11182}, 2026.

\bibitem[Sun et~al.(2019)Sun, Liu, Wu, Pei, Lin, Ou, and
  Jiang]{sun2019bert4rec}
Fei Sun, Jun Liu, Jian Wu, Changhua Pei, Xiao Lin, Wenwu Ou, and Peng Jiang.
\newblock {BERT4Rec: Sequential Recommendation with Bidirectional Encoder
  Representations from Transformer}.
\newblock In \emph{Proceedings of the 28th ACM International Conference on
  Information and Knowledge Management}, pages 1441--1450, 2019.

\bibitem[Templeton et~al.(2024)Templeton, Conerly, Marcus, Lindsey, Bricken,
  Chen, Pearce, Citro, Ameisen, Jones, Cunningham, Turner, McDougall,
  MacDiarmid, Freeman, Sumers, Rees, Batson, Jermyn, Carter, Olah, and
  Henighan]{templeton2024scaling}
Adly Templeton, Tom Conerly, Jonathan Marcus, Jack Lindsey, Trenton Bricken,
  Brian Chen, Adam Pearce, Craig Citro, Emmanuel Ameisen, Andy Jones, Hoagy
  Cunningham, Nicholas~L. Turner, Callum McDougall, Monte MacDiarmid, C.~Daniel
  Freeman, Theodore~R. Sumers, Edward Rees, Joshua Batson, Adam Jermyn, Shan
  Carter, Chris Olah, and Tom Henighan.
\newblock Scaling monosemanticity: Extracting interpretable features from
  {Claude} 3 {Sonnet}.
\newblock \emph{Transformer Circuits Thread}, 2024.
\newblock URL
  \url{https://transformer-circuits.pub/2024/scaling-monosemanticity}.

\bibitem[Wang et~al.(2025)Wang, Zhao, Wang, Zhang, Li, Yu, Wang, Zhou, Yin, Li,
  Guo, and Zhao]{wang2025embeddingrecommendersystemssurvey}
Maolin Wang, Xinjian Zhao, Wanyu Wang, Sheng Zhang, Jiansheng Li, Bowen Yu,
  Binhao Wang, Shucheng Zhou, Dawei Yin, Qing Li, Ruocheng Guo, and Xiangyu
  Zhao.
\newblock Embedding in recommender systems: A survey, 2025.
\newblock URL \url{https://arxiv.org/abs/2310.18608}.

\bibitem[Wang et~al.(2020)Wang, Tang, Lei, Song, Wang, and
  Zhang]{wang2020disenhan}
Yifan Wang, Suyao Tang, Yuntong Lei, Weiping Song, Sheng Wang, and Ming Zhang.
\newblock {DisenHAN}: Disentangled heterogeneous graph attention network for
  recommendation.
\newblock In \emph{Proceedings of the 29th ACM International Conference on
  Information and Knowledge Management (CIKM)}, 2020.

\bibitem[Wei et~al.(2024)Wei, Zelditch, Chen, Ribeiro, Tay, Elizondo, Selvaraj,
  Gupta, and De~Almeida]{wei2024noah}
Changshuai Wei, Benjamin Zelditch, Joyce Chen, Andre Assuncao Silva~T Ribeiro,
  Jingyi~Kenneth Tay, Borja~Ocejo Elizondo, Keerthi Selvaraj, Aman Gupta, and
  Licurgo~Benemann De~Almeida.
\newblock Neural optimization with adaptive heuristics for intelligent
  marketing system.
\newblock In \emph{Proceedings of the 30th ACM SIGKDD Conference on Knowledge
  Discovery and Data Mining}, 2024.

\bibitem[Xiao et~al.(2023)Xiao, Liu, Zhang, and Muennighoff]{bge}
Shitao Xiao, Zheng Liu, Peitian Zhang, and Niklas Muennighoff.
\newblock C-pack: Packaged resources to advance general {C}hinese embedding.
\newblock \emph{arXiv preprint arXiv:2309.07597}, 2023.

\bibitem[Ye et~al.(2024)Ye, O'Neill, Wu, and Iyer]{ye2024steering}
Christine Ye, Charles O'Neill, John~F Wu, and Kartheik~G. Iyer.
\newblock Steering semantic search with interpretable features from sparse
  autoencoders.
\newblock In \emph{MINT: Foundation Model Interventions}, 2024.
\newblock URL \url{https://openreview.net/forum?id=oacksuh7Tu}.

\bibitem[Zaigrajew et~al.(2025)Zaigrajew, Baniecki, and
  Biecek]{zaigrajew2025interpreting}
Vladimir Zaigrajew, Hubert Baniecki, and Przemys{\l}aw Biecek.
\newblock Interpreting {CLIP} with hierarchical sparse autoencoders.
\newblock In \emph{Proceedings of the 42nd International Conference on Machine
  Learning (ICML)}, 2025.

\bibitem[Zheng et~al.(2021)Zheng, Gao, Li, He, Li, and
  Jin]{zheng2021disentangling}
Yu~Zheng, Chen Gao, Xiang Li, Xiangnan He, Yong Li, and Depeng Jin.
\newblock Disentangling user interest and conformity for recommendation with
  causal embedding.
\newblock In \emph{Proceedings of the Web Conference 2021 (WWW)}, pages
  2980--2991, 2021.

\bibitem[Zhu et~al.(2021)Zhu, Liu, Xie, Zhuang, Hao, Ge, Zhang, Lin, and
  Cao]{zhu2021metaheac}
Yongchun Zhu, Yudan Liu, Ruobing Xie, Fuzhen Zhuang, Xiaobo Hao, Kaikai Ge,
  Xu~Zhang, Leyu Lin, and Juan Cao.
\newblock Learning to expand audience via meta hybrid experts and critics for
  recommendation and advertising.
\newblock In \emph{Proceedings of the 27th ACM SIGKDD International Conference
  on Knowledge Discovery \& Data Mining}, 2021.

\end{thebibliography}

\newpage
\appendix
\FloatBarrier
\section{Synthetic data experiments} \label{app:additionalsynthetic}
This appendix comprises additional details regarding the synthetic data experiments of Section~\ref{sec:synthetic}.

\subsection{Setup}
In this experiment, we generate synthetic user--product interactions based on ground truth behavioral modes and test the extent to which various methods can recover them from the embeddings alone. Each ground-truth behavioral mode $\mathcal{P}_k$ is defined by a name, description, product categories, and example products, with an occurrence frequency $\pi_k$ governing its prevalence. We prompted Claude Opus 4.6 to generate $K{=}75$ such modes spanning five domains (Technology \& Gaming, Fitness \& Outdoors, Home \& Kitchen, Lifestyle \& Hobbies, Family \& Personal Care). The resulting set of behavioral modes is shown in Table~\ref{tab:modes}. With the behavioral modes defined, we generate 100,000 synthetic user shopping histories. For each user $\mathcal{U}$, we sample a number of active behavioral modes $n_\mathcal{U}\sim \mathrm{Pois}(\lambda=2)+2$, clipped so that $n_\mathcal{U}\in[2,6]$. The specific $n_\mathcal{U}$ modes are chosen randomly from the set of all behavioral modes with probability proportional to the popularities $\pi_k$, and their weights are drawn independently from $\mathrm{LogNormal}(0, 0.5)$ and normalized to sum to one. We inject noise into the purchase history with weight sampled uniformly from $[0.05,0.2]$. The total number of orders per user is drawn from a log-normal distribution with underlying-normal parameters $\mu=4$ and $\sigma=0.8$ (median $e^4 \approx 55$ interactions), clipped to $[10, 300]$. We allocate the corresponding fraction of interactions to each mode weight plus the noise weight, requiring at least one interaction per active behavioral mode. The generation token budget truncates the longest histories, so realized histories have at most 108 orders (see Appendix~\ref{app:synthetic-stats}).

To generate synthetic user embeddings, we then pass the 100,000 synthetic user profiles described above to Claude Sonnet 4 to generate a corresponding product interaction history for each user. The history generated consists of product names related to each represented behavioral mode. For noise, we provide a list of example product categories and have the model inject similar products. Finally, the user embedding is generated either by embedding each of the product names in a user's history with a sentence embedding model, \texttt{all-mpnet-base-v2}, and taking their mean, or by mean-pooling the SASRec encoder final layer activations across tokens (see Appendix~\ref{app:transformer-synthetic}). Histories parsed successfully for 99{,}832 users, with 5{,}803{,}030 interactions in total. This procedure results in a dense embedding for each user that is inherently a superposition of underlying behavioral modes. For the product name-based embeddings, we $L_2$-normalize and standardize these embeddings, project them onto random directions in $\mathbb{R}^m$ following~\citep{random} to vary the amount of superposition, and then standardize and $L_2$-normalize the projection to arrive at the final representation. For SASRec, details are in Appendix~\ref{app:transformer-synthetic}.

\subsection{Data statistics}\label{app:synthetic-stats}
\begin{figure}[htbp]
\centering
\includegraphics[width=\textwidth]{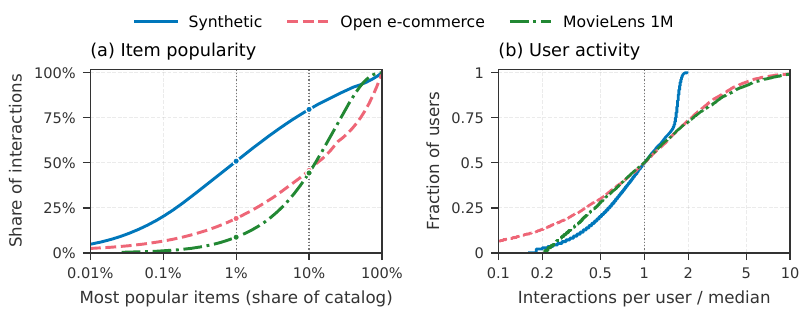}
\caption{Item popularity and user activity for the synthetic data and both real-world datasets. (a) Share of interactions carried by the most popular items, with dots at the top 1\% and 10\% of items (cf.\ Table~\ref{tab:item-popularity}). (b) Interactions per user divided by each dataset's median. Synthetic histories are bounded at about 100 interactions by the generation budget, which truncates the upper tail.}
\label{fig:distributions}
\end{figure}

How closely does our synthetic data reflect properties of real-world user interaction datasets? Table~\ref{tab:item-popularity} and Figure~\ref{fig:distributions} compare item popularity and user activity in the synthetic data with both real datasets considered in our study. The synthetic data is more head-concentrated than either real dataset, with the top 1\% of items carrying 51\% of interactions, and its fraction of singleton items lies between those of Open e-commerce and MovieLens. Below the median, synthetic user activity resembles MovieLens, while Open e-commerce has more low-activity users (p50/p10 activity ratio 2.89, vs.\ 3.56 for MovieLens and 6.63 for Open e-commerce). Above the median, the generation budget truncates the synthetic upper tail near twice the median. The synthetic data therefore reproduces the short head and long tail of item popularity that characterize real-world user interaction data. This implies the recovery results of Section~\ref{sec:synthetic} do not depend on unrealistically uniform item popularity. The synthetic data distribution differs from the real-world examples considered primarily by lacking very heavy users.

\begin{table}[htbp]
\centering
\small
\caption{Item popularity in the synthetic data and both real-world datasets. Columns give the number of items, the mean number of interactions per item, the share of interactions carried by the most popular 1\% and 10\% of items, and the fraction of items that appear exactly once.}
\label{tab:item-popularity}
\begin{tabular}{lrrccc}
\toprule
Dataset & Items & Interactions per item & Top 1\% & Top 10\% & Singletons \\
\midrule
Synthetic & 778{,}639 & 7.5 & 50.9\% & 79.7\% & 62.3\% \\
Open e-commerce (ASIN) & 939{,}082 & 2.0 & 19.1\% & 45.8\% & 75.6\% \\
MovieLens 1M & 3{,}706 & 270 & 8.7\% & 44.4\% & 3.1\% \\
\bottomrule
\end{tabular}
\end{table}

\begin{table}[t]
\centering
\caption{Complete taxonomy of the 75 synthetic behavioral modes used as ground truth for the synthetic experiments of Section~\ref{sec:synthetic}. Each behavioral mode was generated with a name, description, product categories, and example products (not shown for compactness); only the name and domain grouping are displayed here.}
\label{tab:modes}
\small
\begin{tabular}{@{}rll@{}}
\toprule
\textbf{ID} & \textbf{Behavioral mode} & \textbf{Domain} \\
\midrule
\multicolumn{3}{l}{\textit{Technology \& Gaming}} \\
1 & Tech Enthusiast & Tech/Electronics \\
2 & Mobile Gamer & Gaming \\
3 & PC Gamer & Gaming/Hardware \\
4 & Smart Home Builder & IoT/Automation \\
5 & Audiophile & Audio \\
6 & Console Collector & Gaming/Collectibles \\
7 & Photography Hobbyist & Cameras/Lenses \\
8 & Drone Pilot & Drones/Outdoor \\
9 & VR Enthusiast & VR/Gaming \\
10 & Streaming Creator & Audio/Video \\
11 & 3D Printing Maker & Manufacturing \\
12 & Mechanical Keyboard Builder & Keyboards/DIY \\
13 & Home Theater Buff & Audio/Video \\
14 & Tablet Power User & Tablets/Productivity \\
15 & Network Admin Hobbyist & Networking \\
\midrule
\multicolumn{3}{l}{\textit{Fitness \& Outdoors}} \\
16 & Gym Regular & Fitness/Nutrition \\
17 & Marathon Runner & Running/Training \\
18 & Yoga Practitioner & Yoga/Wellness \\
19 & Cycling Enthusiast & Cycling/Outdoor \\
20 & Hiking Adventurer & Hiking/Camping \\
21 & Rock Climber & Climbing/Safety \\
22 & Camping Enthusiast & Camping/Survival \\
23 & Swimmer & Swimming/Fitness \\
24 & CrossFit Athlete & CrossFit/Training \\
25 & Skier/Snowboarder & Winter Sports \\
26 & Tennis Player & Tennis/Sports \\
27 & Golfer & Golf/Sports \\
28 & Kayaker & Water Sports \\
29 & Home Gym Builder & Home Fitness \\
30 & Martial Artist & Martial Arts \\
\midrule
\multicolumn{3}{l}{\textit{Home \& Kitchen}} \\
31 & Home Chef & Cookware/Kitchen \\
32 & Baking Enthusiast & Baking/Tools \\
33 & Coffee Connoisseur & Coffee/Beverages \\
34 & Meal Prepper & Meal Prep/Storage \\
35 & Grilling Master & Grilling/BBQ \\
36 & Wine Enthusiast & Wine/Accessories \\
37 & Home Bartender & Bar/Cocktails \\
38 & Tea Lover & Tea/Beverages \\
39 & Fermentation Hobbyist & Fermentation/DIY \\
40 & Instant Pot User & Appliances \\
41 & Air Fryer Fan & Appliances/Health \\
42 & Sous Vide Cook & Precision Cooking \\
43 & Indoor Gardener & Hydroponics/Plants \\
44 & Kitchen Gadget Collector & Gadgets/Tools \\
45 & Minimalist Chef & Quality Essentials \\
\bottomrule
\end{tabular}
\end{table}

\begin{table}[t]
\centering
\caption{Complete taxonomy of the 75 synthetic behavioral modes used as ground truth (continued).}
\small
\begin{tabular}{@{}rll@{}}
\toprule
\textbf{ID} & \textbf{Behavioral mode} & \textbf{Domain} \\
\midrule
\multicolumn{3}{l}{\textit{Lifestyle \& Hobbies}} \\
46 & Book Lover & Books/Literature \\
47 & Board Game Enthusiast & Gaming/Tabletop \\
48 & Puzzle Solver & Puzzles/Games \\
49 & Craft Enthusiast & Crafts/DIY \\
50 & Knitter/Crocheter & Yarn/Crafts \\
51 & Painting Artist & Art Supplies \\
52 & Model Builder & Model Kits \\
53 & Gardening Hobbyist & Gardening/Outdoor \\
54 & Aquarium Keeper & Aquarium/Fish \\
55 & Vinyl Collector & Music/Audio \\
56 & Journaling Enthusiast & Stationery/Writing \\
57 & Candle Maker & Candle Making/DIY \\
58 & Birdwatcher & Birdwatching/Nature \\
59 & Astronomy Hobbyist & Astronomy/Optics \\
60 & Miniature Painter & Miniatures/Gaming \\
\midrule
\multicolumn{3}{l}{\textit{Family \& Personal Care}} \\
61 & New Parent & Baby/Childcare \\
62 & Pet Parent -- Dog & Pet Supplies \\
63 & Pet Parent -- Cat & Pet Supplies \\
64 & Skincare Enthusiast & Skincare/Beauty \\
65 & Makeup Enthusiast & Cosmetics/Beauty \\
66 & Hair Care Devotee & Hair Care/Styling \\
67 & Fragrance Collector & Perfume/Luxury \\
68 & Natural Beauty Advocate & Organic/Eco-Friendly \\
69 & Nail Art Enthusiast & Nail Care/Beauty \\
70 & Wellness Seeker & Wellness/Health \\
71 & Dental Care Focused & Dental/Health \\
72 & Men's Grooming & Grooming/Skincare \\
73 & Jewelry Lover & Jewelry/Fashion \\
74 & Fashion Shopper & Clothing/Accessories \\
75 & Minimalist Wardrobe & Quality Basics \\
\bottomrule
\end{tabular}
\end{table}

\subsection{SAE and other baselines}
For the SAE, we take the vanilla ReLU version described in Eq.~\eqref{eq:sae}, with an adaptive $\lambda$ sparsity penalty to match the ground truth $L_0$. Here, $\lambda$ is warmed up for 10 epochs, then $\log\lambda$ is adjusted each epoch by an amount proportional to the difference between current and target $L_0$. The SAE has $d_\text{hidden} = 300$ and is trained for 50 epochs with batch size 256 and learning rate $10^{-3}$.

We consider six additional baselines: random projections onto 300 unit vectors (the SAE's width) passed through a ReLU, sparsified versions of PCA and FastICA, K-Means encoding (cosine similarity to K-Means centroids passed through a ReLU and thresholded to match the ground-truth $L_0$, following~\citep{coates2012learning}), the soft-frozen decoder SAE of~\citep{korznikov2026sanitycheckssparseautoencoders}, and NMF on shifted (non-negative) embeddings. K-Means encoding uses 75 components matching the ground-truth mode count, and PCA, ICA, and NMF use $\min(75, m)$. The soft-frozen decoder uses identical hyperparameters to the SAE. To ensure fair comparison, we sparsify all non-SAE methods except NMF to match the ground-truth mean $L_0$ by zeroing activations below a calibrated threshold. NMF retains its natural sparsity, as we find additional thresholding degrades its performance. Note that every time we report results on these methods, they correspond to the sparsified versions.

\subsection{Mode recovery at other cosine thresholds}\label{app:thresholds}
The conclusions of Figure~\ref{fig:comparison_synthetic} do not hinge on the 0.7 cosine cutoff for mode recovery. Repeating the evaluation at thresholds from 0.5 to 0.8, the SAE recovers more modes than FastICA, PCA, and NMF at every threshold and every $m$, and more than every classical baseline in Figure~\ref{fig:comparison_synthetic} for $m \geq 32$. At $m \leq 24$, K-Means encoding matches the SAE at 0.7 ($m = 16$) and exceeds it at 0.8, but its recovery does not improve with dimension ($0.69$ to $0.74$ at 0.8 across all $m$), whereas the SAE's rises from $0.47$ at $m = 16$ to $0.98$ at $m = 128$ and overtakes it from $m = 32$. K-Means centroids thus capture a fixed fraction of the modes regardless of how much room the embedding leaves to separate them, while the SAE utilizes the additional dimensions.

\subsection{Transformer embeddings of the synthetic data}\label{app:transformer-synthetic}
To test recovery when the embedding is produced by a trained sequence encoder, we train SASRec with the configuration of Appendix~\ref{app:movielens-sweep} ($d_\text{model}{=}256$, 4 heads, 2 layers, $\texttt{max\_len}{=}200$, dropout $0.3$, learning rate $5{\times}10^{-4}$) on the synthetic interaction sequences. Items are normalized product names that appear at least 5 times (119{,}826 items covering 83.9\% of interactions, 99{,}213 users), and each user's history is shuffled, since the generative model draws interactions independently. The encoder reaches a test HR@10 of 0.93. User embeddings are the mean of final-layer hidden states, standardized and $L_2$-normalized, and the SAE (with $d_\text{hidden}{=}300$) and baselines are run exactly as in Section~\ref{sec:synthetic}. Items are not mode-exclusive: 15 to 20\% of interaction mass falls on noise items bought across modes, and the mass-weighted mode purity over all items is 0.74 (1.0 would be perfectly exclusive).

Table~\ref{tab:transformer-synthetic} reports recovery at $d{=}256$. A linear readout still recovers most of the mode information (oracle $0.78$, vs.\ $0.85$ for the mean-pooled product embeddings at $m{=}256$), so the encoder preserves the modes, and the SAE recovers them better than the classical baselines. The SAE leads on code--GT alignment and recovers 73 of 75 modes at cosine ${\geq}\,0.7$ (97 to 100\% across repeated SAE trainings on the same embeddings), while the classical baselines lose much more relative to the mean-pooled case, FastICA falling from $95\%$ to $59\%$ recovery. FastICA keeps the highest per-feature correlation, as in Section~\ref{sec:synthetic}, though merging raises the SAE's from $0.28$ to $0.54$. LDA, fit to the raw user--item counts with 75 topics and sparsified to the same $L_0$, recovers real structure but trails the SAE on both metrics it can support, consistent with the claim in Section~\ref{sec:related} that embeddings carry more recoverable mode structure than raw counts. Doubling its training iterations changes its alignment by less than $0.01$.

\begin{table}[htbp]
\centering
\small
\caption{Mode recovery on mean-pooled SASRec embeddings ($d{=}256$) of the synthetic data, all methods at matched $L_0$. Note: LDA is fit to raw user--item counts and has no decoder directions.}
\label{tab:transformer-synthetic}
\begin{tabular}{lccc}
\toprule
Method & Code--GT alignment & Median per-feature $r$ & Modes recovered (cos ${\geq}\,0.7$) \\
\midrule
Oracle (linear ceiling) & 0.78 & --- & --- \\
SAE & \textbf{0.58} & 0.28 & \textbf{0.97} \\
SAE (merged) & 0.54 & 0.54 & 0.95 \\
FastICA & 0.53 & \textbf{0.62} & 0.59 \\
K-Means encoding & 0.45 & 0.30 & 0.52 \\
PCA & 0.25 & 0.17 & 0.00 \\
LDA (raw counts) & 0.37 & 0.27 & --- \\
\bottomrule
\end{tabular}
\end{table}

The same picture holds under further compression and when the encoder itself must superpose the modes. Randomly projecting the $d{=}256$ embeddings to $m = 16$ to $128$ (5 projections each), the SAE recovers more modes than PCA, FastICA, and K-Means encoding at every $m$, although FastICA attains higher alignment for $m \geq 48$. An encoder trained at $d_\text{model}{=}64 < K$ (for which the \emph{encoder} itself is forcing the superposition) has a lower ceiling (oracle $0.58$), yet the SAE still has the best alignment ($0.41$, vs.\ $0.36$ for FastICA) and recovers $97\%$ of modes at cosine ${\geq}\,0.5$ and $71\%$ at ${\geq}\,0.7$, while FastICA recovers none at either threshold. Finally, Table~\ref{tab:k-sensitivity} varies the number of underlying modes. Restricting the population to users whose active modes all lie among the $K'$ most popular modes gives exact $K'$-mode ground truth without having to regenerate the dataset. Across the number of modes considered, the SAE (with $d_\text{hidden} = 4K'$) captures a nearly constant $73$ to $75\%$ of the oracle ceiling while matching or exceeding the best classical baseline at every $K'$. The restriction selects users with fewer and more concentrated modes, which is why we report the oracle-normalized ratio, and $K' = 10$ leaves too few users (2{,}043) to evaluate.

\begin{table}[htbp]
\centering
\small
\caption{Sensitivity to the number of modes $K'$ on the $d{=}256$ transformer embeddings. The columns show the number of corresponding users along with code--GT alignment for the oracle, the SAE, and the best of (PCA, FastICA, K-Means encoding) at each $K'$.}
\label{tab:k-sensitivity}
\begin{tabular}{rrcccc}
\toprule
$K'$ & Users & Oracle & SAE & Best classical & SAE / oracle \\
\midrule
15 & 4{,}591 & 0.943 & 0.695 & 0.667 & 0.737 \\
20 & 8{,}394 & 0.931 & 0.683 & 0.598 & 0.734 \\
25 & 13{,}245 & 0.918 & 0.679 & 0.643 & 0.740 \\
30 & 19{,}880 & 0.897 & 0.651 & 0.631 & 0.725 \\
40 & 36{,}737 & 0.871 & 0.642 & 0.623 & 0.737 \\
50 & 58{,}248 & 0.838 & 0.619 & 0.617 & 0.739 \\
60 & 79{,}007 & 0.809 & 0.604 & 0.572 & 0.747 \\
75 & 99{,}213 & 0.779 & 0.581 & 0.526 & 0.746 \\
\bottomrule
\end{tabular}
\end{table}

\FloatBarrier
\section{Open e-commerce additional details} \label{app:ecommerce}

In this appendix, we provide additional details regarding our experiments on the Open e-commerce dataset.

\subsection{Embedding generation}

For these experiments, the SAE is trained directly on the SVD embeddings described in Section~\ref{sec:open-ecommerce}, so no text input enters the decomposition pipeline at any point. We select the SAE operating point following the decoder-orthogonality criterion of~\citep{chanin2025sparse}, giving $d_\text{hidden}{=}512$ ($4{\times}$ overcomplete), $\lambda{=}5\times 10^{-4}$, 25 epochs, and mean $L_0{\approx}23$. All 512 latents remain live after training (no dead features). We use Claude Sonnet~4.6 to auto-interpret each latent, passing 10 high-activating positives and 10 random negatives and asking the model to identify the behavioral pattern present in the positives and absent in the negatives, producing a short feature name (e.g., ``Premium\_Dog\_Food\_Buyers'') and one-sentence description. Full prompts are in Appendix~\ref{app:prompts}. Reported retrieval numbers are averaged over 5 SAE training seeds with the same operating point, and cross-seed paired Wilcoxon tests are reported in Appendix~\ref{app:stats}.

\subsection{Tasks and ground truth for evaluation}
\label{app:ecommerce-tasks}

To evaluate mode-based audience retrieval, we consider 6 \emph{category retrieval} tasks spanning base rates 8--54\% (pet owner, coffee enthusiast, gardener, board gamer, baby products, musician), and 2 \emph{life-event retrieval} tasks from the companion survey (new parent in 2021, moved residence in 2021). Other survey life events (e.g., divorced, changed jobs) either had insufficient positive counts or did not correlate with category-level behavior changes, and are excluded. Early in experimentation we considered two additional category tasks (Fisher, Photographer) for which the retrieval pipeline's default gating settings did not allow any features through for at least one method. Per-method retuning would eliminate the behavior, but we instead fixed the operating point across methods for apples-to-apples comparison of feature and interpretation quality. For category tasks (Table~\ref{tab:ecommerce-tasks}), positives are defined as users who purchased in the corresponding target product categories in post-2021 windows. For life-event tasks, positives are users who self-reported the event in 2021. For category tasks, every method ranks users using only their pre-2021 windows, so these tasks test whether past behavior predicts future category purchases. For life-event tasks, every method uses windows from 2021 onward, where the event's behavioral shift appears.

\begin{table}[htbp]
\centering
\small
\caption{Open e-commerce retrieval tasks. $n_\text{pos}$ is the number of users satisfying the positive definition out of the evaluated user set for each task block (4085 users with at least one post-2021 window for category retrieval; 4373 users with a 2021 survey response for life-event retrieval).}
\label{tab:ecommerce-tasks}
\begin{tabular}{llcc}
\toprule
Task type & Task name & $n_\text{pos}$ & Base rate \\
\midrule
\multirow{6}{*}{Category retrieval}
  & Musician          & 345  & 8.4\%  \\
  & Baby products     & 839  & 20.5\% \\
  & Board gamer       & 1421 & 34.8\% \\
  & Coffee enthusiast & 1482 & 36.3\% \\
  & Gardener          & 1539 & 37.7\% \\
  & Pet owner         & 2196 & 53.8\% \\
\midrule
\multirow{2}{*}{Life-event retrieval}
  & New parent (2021)       & 226 & 5.2\%  \\
  & Moved residence (2021)  & 942 & 21.5\% \\
\bottomrule
\end{tabular}
\end{table}

\subsection{Retrieval strategies}

Here we detail how we retrieve users in our experiments for each mode recovery method.

\emph{SAE + NL} (main-text primary) presents all feature interpretations to an LLM alongside a natural-language audience query, which scores each feature's relevance on a 0--10 scale. We then retain features scoring $\geq 5$ and rank users by the weighted sum of their activations on the retained features. This strategy is fully zero-shot: it sees only the NL query and the feature interpretations.

\emph{K-Means encoding + NL} uses the same scoring and ranking pipeline, substituting K-Means-encoding features for SAE features.

The \emph{text-embedding baseline} requires no sparse decomposition. For each user we embed the NL query with \texttt{all-MiniLM-L6-v2} and rank users by cosine similarity to a per-user text vector. The user vector is the \emph{category-pooled} variant: a TF-IDF-weighted mean of sentence-transformer embeddings over the user's purchased category names (with underscores replaced by spaces and lowercased so the encoder receives natural-language phrases). Note that the SAE, K-Means encoding, and ICA methods consume exactly the same TF-IDF-weighted category signal, so any retrieval advantage over this baseline is attributable to the decomposition itself rather than to extra text the baseline did not see.

\emph{SAE + correlation} (reported here for completeness only; not zero-shot) selects the 20 SAE features most correlated with target-category purchases in pre-2021 data, then ranks users by their weighted activations on those features. This requires access to the target category list for each task and is therefore not a zero-shot method. We report its numbers as a reference point for how much signal the SAE features carry when supervised feature selection is applied. On the category tasks at matched $L_0$, it reaches mean Lift@$n_\text{pos}$ of ${\approx}1.9\times$ and mean AUROC of ${\approx}0.72$, roughly matching SAE+NL. Zero-shot feature selection therefore loses little relative to supervised selection from the same dictionary.

\subsection{Open-weight interpretation model}\label{app:open-model}
In our primary experiments, we chose Claude Sonnet~4.6 for interpretation because it was among the strongest models at its scale and was available for large-scale batch inference on our platform. To test whether the retrieval results depend on the interpretation model, we re-name the features of one Open e-commerce seed with the open-weight Llama~3.3~70B Instruct (temperature 0), holding everything else fixed: the SAE and baseline codes, the sampled users and prompts, and the Claude Haiku~4.5 query-time scoring are unchanged. The feature names produced by Llama closely track those from Sonnet. The median sentence-embedding cosine similarity between the two models' name-and-description strings for the same feature is 0.64 to 0.69 across methods, against 0.29 to 0.35 for different features. We found that Llama assigns nearly constant confidence scores ($0.92 \pm 0.01$), so the 0.6 confidence filter passes every feature, and we therefore also evaluate Llama's names paired with Sonnet's confidence scores. In both variants the SAE remains ahead of every baseline on mean Lift in both task blocks, and its own Lift stays within 6\% of the Sonnet-named value on at least 6 of the 8 tasks. Its margins over FastICA and K-Means encoding narrow, however. With Llama's names, the SAE's category-task AUROC lead over K-Means encoding falls from 0.031 to under 0.01, and its life-event Lift lead over FastICA from 0.32 to about 0.03. For scale, re-running query-time scoring on the Sonnet names alone changes Lift by up to 0.012 on category tasks and 0.09 on life-event tasks, since the scoring is not fully deterministic. The direction of the comparison therefore does not depend on the interpretation model, but its size partly does. Llama's names lead more features per query to pass the retrieval relevance threshold for every method (21 vs.\ 14 for the SAE), which raises the baselines' coverage (K-Means encoding on category tasks, 0.61 to 0.75) while the SAE's barely changes. Part of the SAE's margin under Sonnet's feature interpretations thus comes from how features are named and selected rather than from the decomposition alone. Note that this replication used a single seed and no significance test.

\FloatBarrier
\section{Additional MovieLens experimental details} \label{app:movielens}

This appendix contains additional information relevant for the MovieLens transformer-based experiments described in the main text.

\subsection{Model settings}\label{app:movielens-sweep}

\paragraph{Transformer recommenders.} In our experiments, we train SASRec~\citep{kang2018self} and BERT4Rec~\citep{sun2019bert4rec} with $d_\text{model}{=}256$, 4 heads, 2 layers, $\texttt{max\_len}{=}200$, dropout $0.3$, learning rate $5{\times}10^{-4}$. The encoders are trained on chronologically ordered interactions. User embeddings are the mean of non-padding final-layer hidden states.

\paragraph{SAE operating point.} We selected the SAE operating point reported in Section~\ref{sec:movielens} using the decoder-orthogonality criterion of~\citep{chanin2025sparse}. Specifically, we sweep $\lambda$ at a fixed width and pick the elbow of mean pairwise decoder cosine similarity ($c_\text{dec}$). Among widths satisfying this elbow condition, we choose the smallest width that maintains a high absolute count of features with sufficient statistical support for reliable auto-interpretation (30--500 activating users, ``eligible'' in what follows). Our final choice is $d_\text{hidden}{=}1024$ with $\lambda{=}1.98{\times}10^{-3}$. The same $\lambda$ is used for both encoders. Measured $L_0$ differs only slightly (${\approx}80$ on SASRec, ${\approx}83$ on BERT4Rec), giving a direct architecture-level comparison at a matched sparsity target.

\paragraph{Width and sparsity.}
To illustrate our SAE width and sparsity selection method, Table~\ref{tab:movielens-sweep} reports representative rows of the sweep on SASRec embeddings at $d_\text{hidden}{=}1024$. $c_\text{dec}$ flattens for $L_0 \in [65, 100]$, indicating an elbow beyond which additional sparsity pressure does not further improve decoder orthogonality. Within this band, $L_0{\approx}80$ maximizes eligible feature count. The same qualitative picture holds for BERT4Rec embeddings. Widths below 1024 ($d{\in}\{256, 512\}$) cap eligible-feature yield at under 30\% and were excluded. The $d{=}2048$ configuration exhibited high dead-feature counts without improving either metric and so was also disfavored.

\begin{table}[htbp]
\centering
\small
\caption{Width-and-sparsity sweep on MovieLens SASRec embeddings at $d_\text{hidden}{=}1024$. The selected operating point (bold) sits at the $c_\text{dec}$ elbow with the highest eligible-feature count. ``Eligible'' here counts features with 30--500 activating users.}
\label{tab:movielens-sweep}
\begin{tabular}{lccccc}
\toprule
$\lambda$ & mean $L_0$ & $c_\text{dec}$ & \% eligible & n eligible & recon.\ MSE \\
\midrule
$5.70{\times}10^{-3}$ & 68.7 & 0.095 & 24.5\% & 251 & 0.060 \\
$3.62{\times}10^{-3}$ & 65.9 & 0.080 & 24.6\% & 252 & 0.044 \\
$\boldsymbol{1.98{\times}10^{-3}}$ & \textbf{82.8} & \textbf{0.081} & \textbf{37.1\%} & \textbf{380} & \textbf{0.026} \\
$1.46{\times}10^{-3}$ & 102.3 & 0.083 & 43.6\% & 446 & 0.021 \\
$8.30{\times}10^{-4}$ & 147.7 & 0.086 & 38.0\% & 389 & 0.014 \\
\bottomrule
\end{tabular}
\end{table}

\paragraph{Resulting decoder-cosine distributions.} Pairwise decoder vector cosine similarities are centered near zero for the SAE on both encoders (mean $0.0001$ on SASRec, $0.0008$ on BERT4Rec), confirming the near-orthogonal geometry claimed in Section~\ref{sec:movielens}. K-Means encoding's centroids are natively denser (mean pairwise cosine $0.29$ on SASRec, $0.17$ on BERT4Rec).

\paragraph{Training length.}
Table~\ref{tab:training-length} reports a training-length sensitivity check at the selected $(d, \lambda)$ configuration. Training past 250 epochs begins to over-sparsify the dictionary: eligible-feature count declines as features slip below the 30-user support threshold, and $c_\text{dec}$ worsens because the dying features leave behind overlapping decoder directions. Shorter training (125 epochs) produces a higher raw eligible count, but inspection of the top features reveals within-theme redundancy (e.g.\ multiple distinct ``classic Western'' latents) that an overcomplete dictionary has not yet deduplicated. We therefore train to 250 epochs for both encoders, which sits at the inflection point between under- and over-training.

\begin{table}[htbp]
\centering
\small
\caption{Training-length sensitivity at $d_\text{hidden}{=}1024$, $\lambda{=}1.98{\times}10^{-3}$ on SASRec embeddings. The selected training length (bold) balances feature count against within-theme redundancy. Metrics at 250 epochs differ from the corresponding row in Table~\ref{tab:movielens-sweep} because the two tables report independent training runs.}
\label{tab:training-length}
\begin{tabular}{lccccc}
\toprule
Epochs & mean $L_0$ & $c_\text{dec}$ & n eligible & n too-rare & recon.\ MSE \\
\midrule
125 & 101.0 & 0.054 & 653 & 0   & 0.091 \\
\textbf{250} & \textbf{80.2} & \textbf{0.059} & \textbf{572} & \textbf{125} & \textbf{0.052} \\
500 & 74.3 & 0.076 & 498 & 165 & 0.033 \\
\bottomrule
\end{tabular}
\end{table}

\subsection{Natural-language audience queries and ground truth}\label{app:nl-retrieval}

Table~\ref{tab:nl-queries} lists the 32 natural-language audience queries used in the MovieLens retrieval evaluation, along with the title patterns and optional MovieLens-genre constraint used to resolve each query to a ground-truth positive set. Ground-truth labels are generated by a deterministic rule: for query $Q$ with title-pattern set $\mathcal{P}_Q$ and optional genre constraint $\mathcal{G}_Q$, a user is labelled positive if their MovieLens watch history contains $\geq 3$ movies whose title matches any pattern in $\mathcal{P}_Q$ (case-insensitive substring match) \emph{and} whose MovieLens-listed genres include every genre in $\mathcal{G}_Q$. Queries yielding fewer than 20 positive users in the evaluated population are dropped from the benchmark. We originally generated 33 queries, and this criterion dropped one: \emph{``viewers who enjoy martial arts and kung fu films''}. The remaining 32 are retained. Queries split into 21 broad audiences (e.g.\ Tarantino fans, Disney animation) and 11 niche audiences (e.g.\ French New Wave, pre-1960 film noir). Draft query strings and candidate title-pattern lists were generated with Claude Opus~4.6 and then curated by the authors against the MovieLens~1M catalog. The final patterns used to produce ground-truth labels are those listed in Table~\ref{tab:nl-queries}. Paired Wilcoxon p-values for per-query SAE-vs-baseline comparisons are consolidated in Appendix~\ref{app:stats}, Table~\ref{tab:stats}.

The text-embedding baseline evaluated alongside SAE and K-Means encoding scores each user by the cosine similarity between the query embedding and the user's mean-pooled watched-title embedding (both computed with \texttt{all-MiniLM-L6-v2}).

\begin{longtable}{@{}p{0.24\linewidth}p{0.43\linewidth}p{0.15\linewidth}r@{}}
\caption{Natural-language audience queries on MovieLens~1M with title-pattern and genre-constraint ground-truth definitions. $n_\text{+}$ is the resulting number of positive users out of 6040.}
\label{tab:nl-queries} \\
\toprule
Query & Title patterns (representative) & Genre & $n_\text{+}$ \\
\midrule
\endfirsthead
\multicolumn{4}{c}{\tablename~\thetable{} (continued)} \\
\toprule
Query & Title patterns (representative) & Genre & $n_\text{+}$ \\
\midrule
\endhead
\midrule
\multicolumn{4}{r}{Continued on next page.} \\
\endfoot
\bottomrule
\endlastfoot
\multicolumn{4}{l}{\textit{Broad audiences (21 queries)}} \\
\midrule
Fans of Quentin Tarantino films & Pulp Fiction, Reservoir Dogs, Jackie Brown, Kill Bill, Four Rooms & --- & 517 \\
Fans of Steven Spielberg adventure and sci-fi films & Raiders of the Lost Ark, Indiana Jones, Jurassic Park, E.T., Close Encounters, \ldots (9 total) & --- & 3099 \\
Fans of classic Disney animated films & Lion King, Little Mermaid, Beauty and the Beast, Aladdin, Snow White, \ldots (21 total) & --- & 1682 \\
Viewers of dark, psychological thrillers & Silence of the Lambs, Se7en, Fight Club, American Psycho, Memento, \ldots (13 total) & --- & 1748 \\
Fans of romantic comedies from the 1990s & Pretty Woman, Sleepless in Seattle, You've Got Mail, Notting Hill, Four Weddings, \ldots (10 total) & Romance/\allowbreak Comedy & 1237 \\
Fans of Stanley Kubrick films & Clockwork Orange, 2001: A Space Odyssey, Shining, Full Metal Jacket, Dr. Strangelove, \ldots (10 total) & --- & 1385 \\
Fans of Star Wars and space opera & Star Wars, Empire Strikes Back, Return of the Jedi, Phantom Menace & --- & 2237 \\
Viewers who enjoy classic film noir and crime dramas & Chinatown, Maltese Falcon, Double Indemnity, Third Man, Sunset Blvd, \ldots (10 total) & --- & 1019 \\
Fans of Monty Python and British comedy & Monty Python and the Holy Grail, Life of Brian, Meaning of Life, Fish Called Wanda, Time Bandits & --- & 758 \\
Viewers who enjoy horror and slasher films & Nightmare on Elm Street, Halloween, Friday the 13th, Texas Chain Saw, Scream, \ldots (14 total) & Horror & 1078 \\
Fans of epic historical war films & Saving Private Ryan, Braveheart, Gladiator, Schindler's List, Platoon, \ldots (11 total) & War & 2228 \\
Fans of quirky independent and arthouse films & Fargo, Big Lebowski, Raising Arizona, Barton Fink, Being John Malkovich, \ldots (13 total) & --- & 4504 \\
Fans of slapstick and physical comedy & Airplane!, Naked Gun, Blazing Saddles, Young Frankenstein, Ace Ventura, \ldots (11 total) & --- & 1887 \\
Fans of Tim Burton's gothic and fantasy films & Edward Scissorhands, Beetlejuice, Batman, Batman Returns, Nightmare Before Christmas, \ldots (8 total) & --- & 1591 \\
Viewers who enjoy coming-of-age teen dramas & Breakfast Club, Sixteen Candles, Ferris Bueller, Stand By Me, Dead Poets Society, \ldots (10 total) & --- & 1545 \\
Fans of classic musicals & Singin' in the Rain, Sound of Music, West Side Story, Wizard of Oz, My Fair Lady, \ldots (10 total) & Musical & 1055 \\
Viewers interested in courtroom and legal dramas & 12 Angry Men, Few Good Men, To Kill a Mockingbird, Verdict, Primal Fear, \ldots (10 total) & --- & 930 \\
Fans of Woody Allen films & Annie Hall, Manhattan, Hannah and Her Sisters, Crimes and Misdemeanors, Sleeper, \ldots (13 total) & --- & 935 \\
Fans of the Coen Brothers films & Fargo, Big Lebowski, Raising Arizona, Barton Fink, Miller's Crossing, \ldots (9 total) & --- & 873 \\
Fans of Martin Scorsese films & Taxi Driver, Raging Bull, GoodFellas, Casino, Mean Streets, \ldots (12 total) & --- & 945 \\
Fans of Alfred Hitchcock films & Psycho, Vertigo, Rear Window, North by Northwest, Birds, \ldots (17 total) & --- & 1301 \\
\midrule
\multicolumn{4}{l}{\textit{Niche audiences (11 queries)}} \\
\midrule
Viewers who enjoy Studio Ghibli and Japanese animation & Princess Mononoke, My Neighbor Totoro, Kiki's Delivery Service, Nausicaä, Castle in the Sky, \ldots (8 total) & --- & 97 \\
Viewers who enjoy documentary films & Hoop Dreams, Roger \& Me, Thin Blue Line, Crumb, Buena Vista Social Club, American Movie & Documentary & 329 \\
Fans of David Lynch films & Eraserhead, Blue Velvet, Wild at Heart, Lost Highway, Mulholland Dr., \ldots (8 total) & --- & 176 \\
Fans of 1970s New Hollywood auteur films & Chinatown, Taxi Driver, Godfather, Godfather: Part II, Nashville, \ldots (16 total) & --- & 1787 \\
Fans of 1980s slasher horror films & Halloween, Friday the 13th, Nightmare on Elm Street, Prom Night, Terror Train, \ldots (11 total) & Horror & 583 \\
Fans of classic 1930s--40s screwball comedies & Bringing Up Baby, It Happened One Night, His Girl Friday, Philadelphia Story, My Man Godfrey, \ldots (14 total) & Comedy & 432 \\
Fans of French New Wave cinema & Breathless, 400 Blows, Jules and Jim, Contempt, Band of Outsiders, \ldots (12 total) & --- & 47 \\
Fans of pre-1960 film noir & Double Indemnity, Maltese Falcon, Out of the Past, Big Sleep, Sunset Blvd, \ldots (15 total) & Film-Noir & 540 \\
Viewers of silent-era cinema & Metropolis, General, Gold Rush, City Lights, Nosferatu, \ldots (11 total) & --- & 230 \\
Fans of classic Japanese samurai cinema & Seven Samurai, Yojimbo, Sanjuro, Throne of Blood, Ran, \ldots (10 total) & --- & 1724 \\
Fans of Italian neorealism and post-war Italian cinema & Bicycle Thief, La Dolce Vita, 8 1/2, La Strada, Umberto D, \ldots (11 total) & --- & 60 \\
\end{longtable}

The paired Wilcoxon p-values for MovieLens retrieval are consolidated in Appendix~\ref{app:stats}, Table~\ref{tab:stats}, alongside the Open e-commerce and audience-characterization tests. The main conclusions for our MovieLens experiments are as follows. The SAE and K-Means encoding are statistically indistinguishable on Lift@$n_\text{pos}$ and AUROC. The SAE wins Coverage by a large and highly significant margin on both encoders. The SAE beats the text-embedding baseline on Lift but is nominally tied on AUROC, consistent with the text baseline's dense-by-construction coverage providing it a full-population comparison set. The AUROC$_\text{scored}$ column of Table~\ref{tab:movielens-main} restricts to users each method actually scored, which leaves the text baseline slightly ahead on this metric.

\subsection{Held-out demographic labels}\label{app:demographics}
Because the MovieLens ground truth is derived from titles, which the auto-interpretation prompt also shows the labeling model, retrieval could in principle reflect LLM title matching rather than the decomposition. As a control, we retrieve against all 14 of MovieLens's self-reported demographic labels (age, gender, and occupation groups), which no stage of the pipeline sees, changing nothing else. The SAE reaches AUROC 0.58 (SASRec) and 0.60 (BERT4Rec) and Lift 2.00 and 2.06, against AUROC 0.54 and Lift 1.28 for the text-embedding baseline (one-sided paired Wilcoxon on AUROC over query and seed, $p \leq 3{\times}10^{-4}$). K-Means encoding also exceeds the text baseline, so the effect reflects sparse decomposition rather than our SAE specifically. If the benchmark measured title matching, both methods would degrade together. Instead, only the method that depends on item text does.

\subsection{Sensitivity to the confidence threshold}\label{app:conf-threshold}
The auto-interpretation confidence filter (${\geq}\,0.6$, cf.\ Section~\ref{sec:recovery}) removes about a third of SAE features on MovieLens but only a tenth of K-Means features, because the LLM assigns the broader K-Means features higher confidence. Table~\ref{tab:conf-threshold} re-runs retrieval at thresholds from 0.0 (no filter) to 0.7. Across this range, the Lift and AUROC differences between the SAE and K-Means encoding stay small, and the SAE maintains its advantage in coverage, which grows with the threshold. The threshold therefore acts as a resolution parameter rather than a tuned setting.

\begin{table}[htbp]
\centering
\small
\caption{Sensitivity of MovieLens retrieval to the auto-interpretation confidence threshold (operating point 0.6). We show results in terms of the difference between SAE and K-Means encoding values. Here, change in Lift is shown as a percentage of the K-Means value, AUROC and AUROC$_\text{scored}$ as absolute differences, and coverage as the raw SAE / K-Means pair. Values are computed over 3 SAE seeds per encoder, a different seed set from Table~\ref{tab:movielens-main}.}
\label{tab:conf-threshold}
\begin{tabular}{llcccc}
\toprule
Encoder & Threshold & $\Delta$Lift & $\Delta$AUROC & $\Delta$AUROC$_\text{scored}$ & Coverage (SAE / K-Means) \\
\midrule
\multirow{4}{*}{SASRec}
 & 0.0 & $+6.3\%$ & $-0.016$ & $-0.004$ & 0.973 / 0.818 \\
 & 0.5 & $+10.8\%$ & $+0.000$ & $+0.011$ & 0.976 / 0.817 \\
 & 0.6 & $+11.2\%$ & $+0.009$ & $+0.018$ & 0.975 / 0.808 \\
 & 0.7 & $+7.4\%$ & $+0.019$ & $+0.019$ & 0.983 / 0.795 \\
\midrule
\multirow{4}{*}{BERT4Rec}
 & 0.0 & $-5.8\%$ & $-0.019$ & $+0.003$ & 0.971 / 0.847 \\
 & 0.5 & $-4.2\%$ & $-0.016$ & $+0.006$ & 0.973 / 0.846 \\
 & 0.6 & $-2.9\%$ & $-0.009$ & $+0.013$ & 0.979 / 0.843 \\
 & 0.7 & $-4.3\%$ & $-0.006$ & $+0.012$ & 0.979 / 0.828 \\
\bottomrule
\end{tabular}
\end{table}

\subsection{Audience characterization}\label{app:audience-characterization}

Here we describe the audience characterization analysis of Section~\ref{sec:audience}. The goal is to describe a given audience by the features most over-represented in that group of users relative to the full population. Since each selected feature is distinctive of the audience, the more semantically diverse the selected set, the more informative the characterization. We consider 20 franchise-level audiences in MovieLens-1M ($\geq 2$ franchise films per user, $\geq 30$ users per audience) and select $k{=}5$ features per audience for each method. A user counts as active on a feature if their activation is in the top $10\%$ of that feature's nonzero activations (matching the auto-interpretation sampling convention). This places SAE and K-Means features, whose activation scales differ, on an equal footing. We rank features by lift, the fraction of the audience activating the feature divided by the fraction of all users activating it in the full population. The K-Means-encoding reference is thresholded to match the SAE's mean $L_0$, following Section~\ref{sec:realdata-setup}.

\paragraph{Top-$k$ diversity.} Per audience, we compute top-$k$ \emph{redundancy} as the median pairwise cosine between BGE-large-en-v1.5~\citep{bge} embeddings of the top-$k$ features' \{feature\_name, feature\_description\} strings (lower cosine = less redundant, more diverse). We use BGE rather than the retrieval-standard MiniLM here because BGE provides finer paraphrase-discrimination required for this metric, which specifically targets near-paraphrases that retrieval-tuned encoders collapse by design. The SAE decomposition yields a significantly less redundant top-5 feature set than K-Means encoding on both encoders (SASRec K-Means $-$ SAE median cosine $\Delta{=}0.066$, $p{=}0.002$; BERT4Rec $\Delta{=}0.032$, $p{=}0.049$; Stouffer $p{\approx}4{\times}10^{-4}$, one-sided paired Wilcoxon with alternative K-Means$>$SAE, across 20 audiences).

\paragraph{Reverse-selector audience coverage.} To probe whether redundancy costs audience coverage, we reverse the feature selector: we rank features by cosine similarity between their \{name, description\} and an LLM-generated audience description, take the top-5, then measure the fraction of each audience's top-activators covered by the selected top-5. The SAE top-5 covers significantly more of the target audience (SASRec $\Delta{=}0.011$, $p{=}0.0003$; BERT4Rec $\Delta{=}0.017$, $p{=}0.0001$; Stouffer $p{\approx}10^{-7}$). The K-Means-encoding reference has higher per-feature lift but narrower per-feature activation distributions, so its redundant top-5 does not translate into better audience coverage. The SAE's L1-sparsity and overcomplete dictionary produce a decomposition in which each selected direction explains a different part of the audience's behavior, and this complementarity is what the coverage measure rewards.

\paragraph{Feature-level redundancy diagnostic.}\label{app:redundancy} The top-$k$ diversity result above is measured on feature names and descriptions. Here we check that it also holds in terms of the features' activations. For each of the 10 curated niche audiences in Appendix~\ref{app:nl-retrieval} with $n_\text{positive}{\leq}500$, we identify SAE and K-Means features whose top-50 activators overlap the audience's positive set (Jaccard${\geq}0.05$) and whose LLM interpretation confidence is ${\geq}0.6$ (matching the retrieval pipeline's filter). Among these features, we compute pairwise activation Pearson correlation and an effective-concept count by agglomerative clustering at correlation threshold $0.5$. The redundancy ratio is features per effective concept. Across the 5 niche audiences where both methods have ${\geq}2$ confidence-passing matched features (documentary, David Lynch, 1930s--40s screwball, French New Wave, Italian neorealism), the SAE has mean activation correlation $0.16$ vs K-Means $0.29$, and a redundancy ratio of $1.00$ vs $2.20$. K-Means features that surface a single audience decompose into $2$--$4$ redundant partitions per effective concept; SAE features are structurally distinct even when multiple features co-surface the same audience.

\paragraph{Open e-commerce feature-level redundancy.}\label{app:redundancy-ecommerce} Applying the same diagnostic to the SVD-on-counts SAE, with category audiences as ground truth and a precision-$\geq 0.70$ match criterion (audience sizes of $300$--$2200$ make Jaccard-based matching saturate at low values), we find the effect becomes more pronounced: the SAE exhibits mean activation correlation $0.09$ vs K-Means $0.33$, with redundancy ratio $1.14$ vs $2.82$ across $4$/$7$ audiences with ${\geq}2$ matched features per method. $93\%$ of active SAE features pass the confidence filter (vs $65\%$ on MovieLens), consistent with the cleaner mode-to-feature alignment predicted for linear embedding functions in Section~\ref{sec:method}.

\subsection{Volume-axis relabeling}\label{app:volume-relabel}

We close this appendix with a description of our volume-axis relabeling method for MovieLens features. A sparse decomposition of sequence-encoder embeddings can produce features that track engagement \emph{intensity} (total films watched, recency of activity) rather than any taste pattern. Because such features fire on users with high activity regardless of content preference, they inflate retrieval lift on any audience query that correlates with engagement volume, such as broad ``fans of Hollywood films'' queries where the positive set is dominated by heavy watchers.

We identify these features via a length-matched preference-free control: for each feature, we compare the content profile of its top activators against a group of non-activating users whose viewing counts match the activators' volume distribution. A feature is classified as \emph{pure volume} if the content profiles are statistically indistinguishable (no genre or era is significantly enriched in the activator group relative to the length-matched control at $p{<}0.05$ after multiple-testing correction). Pure-volume features are relabeled from their auto-interpreted name to a single engagement-axis label (``heavy-watcher engagement axis'') before retrieval. The same detection procedure is applied to SAE and K-Means-encoding codes, ensuring both methods receive the same treatment.

Relabeled engagement features contribute to retrieval only to the extent that the LLM selects them based on semantic match to the query text. The protocol therefore ensures that broad audience queries correlated with viewing volume are not spuriously boosted by engagement features. The features remain available in the dictionary but are interpreted honestly as an engagement axis rather than a preference.

\FloatBarrier
\section{Additional baselines: PCA, ICA on Open e-commerce and MovieLens}\label{app:baselines}

The main text reports the SAE and K-Means encoding results on both real-world datasets, along with FastICA on Open e-commerce (its closest classical competitor) and a text-embedding baseline. For completeness, Table~\ref{tab:baselines-all} reports PCA and FastICA results on both datasets at matched mean $L_0$. PCA and ICA are capped at the input dimension by construction ($d{=}128$ on Open e-commerce, $d{=}256$ on MovieLens), since orthogonal/independent components cannot be overcomplete. SAE and K-Means encoding use the SAE's dictionary width ($d_\text{hidden}{=}512$ for Open e-commerce, $d_\text{hidden}{=}1024$ for MovieLens). Results are averaged across tasks within each dataset.

\begin{table}[htbp]
\centering
\small
\caption{Additional baselines on both real-world datasets. Metrics are averaged across tasks. The SAE significantly exceeds every classical baseline on Lift at matched $L_0$ (cross-seed Wilcoxon, Appendix~\ref{app:stats}). ICA is the SAE's closest competitor on AUROC (tied on life-event AUROC, SAE wins elsewhere) and achieves higher Coverage by construction (denser activations). NMF is excluded as it requires non-negative preprocessing that does not naturally arise in either embedding space and underperforms classical baselines at all $m$ (Section~\ref{sec:synthetic}).}
\label{tab:baselines-all}
\begin{tabular}{llcccc}
\toprule
Dataset & Method & FVU $\downarrow$ & Lift & AUROC & Coverage \\
\midrule
\multirow{2}{*}{Open e-commerce}
  & PCA              & 0.388 & 1.28$\times$ & 0.579 & 0.797 \\
  & ICA              & 0.083 & 1.86$\times$ & 0.695 & 0.770 \\
\midrule
\multirow{2}{*}{MovieLens (SASRec)}
  & PCA  & 0.189         & 1.65          & 0.576 & 0.815 \\
  & ICA  & 0.343         & 1.99          & 0.610 & 0.910 \\
\midrule
\multirow{2}{*}{MovieLens (BERT4Rec)}
  & PCA  & 0.200         & 1.60          & 0.592 & 0.904 \\
  & ICA  & 0.357         & 2.08          & 0.580 & 0.833 \\
\bottomrule
\end{tabular}
\end{table}

\FloatBarrier
\section{Statistical significance tables}\label{app:stats}

Table~\ref{tab:stats} consolidates all Wilcoxon signed-rank tests and Stouffer combinations referenced in the main text. All tests are one-sided with alternative hypothesis SAE $>$ baseline. The direction was pre-specified by the hypothesis before observing the data. For tests where the baseline wins, we still report the one-sided $p$ for SAE $>$ baseline, which is then large, rather than suppressing the comparison.

Open e-commerce tests are paired across 5 seeds $\times$ task (6 category tasks or 2 life-event tasks) as cross-seed paired observations. MovieLens retrieval tests are paired across the 32 corresponding queries, reported separately per encoder as well as Stouffer-combined. Audience-characterization tests are paired across 20 franchise audiences, Stouffer-combined across the two MovieLens encoders.

\begin{table}[htbp]
\centering
\small
\caption{Consolidated paired Wilcoxon one-sided $p$-values. Default alternative H1: SAE $>$ baseline on the named metric. Bold indicates $p{<}0.05$ \emph{in the SAE-wins direction}, and comparisons the baseline wins are reported unbolded. MovieLens retrieval columns report per-encoder and Stouffer-combined $p$; Open e-commerce and audience-characterization sections are single-dataset (e-commerce) or Stouffer-combined across MovieLens encoders (audience). Using Stouffer's combined test across encoders assumes independence; here the two encoders share users and queries, so combined $p$-values should be read as descriptive summaries rather than independence-based combinations (per-encoder columns are the primary quantity). \textbf{Note on pairing:} Open e-commerce and MovieLens SAE-vs-K-Means rows are cross-seed (Open e-commerce: 5 seeds $\times$ task; MovieLens: 2 SAE-training seeds $\times$ 32 queries). MovieLens rows vs. PCA/ICA/text are single-seed paired across 32 queries. $^\dagger$For top-5 redundancy (median pairwise cosine among the top-5 features), smaller is more diverse; alternative hypothesis is H1: K-Means redundancy $>$ SAE redundancy. $^\ddagger$The named comparison wins on this metric, so the one-sided $p$ for SAE $>$ baseline is large.}
\label{tab:stats}
\setlength{\tabcolsep}{3.4pt}
\begin{tabular}{llccc}
\toprule
 & Comparison & \multicolumn{3}{c}{One-sided $p$} \\
\midrule
\multicolumn{5}{l}{\textit{Open e-commerce category retrieval (cross-seed paired Wilcoxon, 5 seeds $\times$ 6 tasks)}} \\
Lift@$n_\text{pos}$  & SAE vs K-Means encoding       & \multicolumn{3}{c}{$\mathbf{p = 0.0003}$} \\
Lift@$n_\text{pos}$  & SAE vs ICA                    & \multicolumn{3}{c}{$\mathbf{p = 0.0002}$} \\
Lift@$n_\text{pos}$  & SAE vs PCA                    & \multicolumn{3}{c}{$\mathbf{p < 10^{-4}}$} \\
AUROC                & SAE vs K-Means encoding       & \multicolumn{3}{c}{$\mathbf{p < 10^{-4}}$} \\
AUROC                & SAE vs ICA                    & \multicolumn{3}{c}{$\mathbf{p = 0.004}$} \\
AUROC                & SAE vs PCA                    & \multicolumn{3}{c}{$\mathbf{p < 10^{-4}}$} \\
AUROC$_\text{scored}$ & SAE vs K-Means encoding      & \multicolumn{3}{c}{$\mathbf{p = 0.009}$} \\
AUROC$_\text{scored}$ & SAE vs ICA                   & \multicolumn{3}{c}{$\mathbf{p = 0.0003}$} \\
AUROC$_\text{scored}$ & SAE vs PCA                   & \multicolumn{3}{c}{$\mathbf{p < 10^{-4}}$} \\
Coverage             & SAE vs K-Means encoding       & \multicolumn{3}{c}{$\mathbf{p < 10^{-4}}$} \\
Coverage             & SAE vs ICA\textsuperscript{$\ddagger$}                    & \multicolumn{3}{c}{$p = 0.91$ (ICA wins)} \\
Coverage             & SAE vs PCA\textsuperscript{$\ddagger$}                    & \multicolumn{3}{c}{$p = 1.0$ (PCA wins)} \\
\midrule
\multicolumn{5}{l}{\textit{Open e-commerce life-event retrieval (cross-seed paired Wilcoxon, 5 seeds $\times$ 2 tasks)}} \\
Lift@$n_\text{pos}$  & SAE vs K-Means / ICA / PCA    & \multicolumn{3}{c}{$\mathbf{p = 0.001}$ (each)} \\
AUROC                & SAE vs K-Means / PCA          & \multicolumn{3}{c}{$\mathbf{p = 0.001}$ (each)} \\
AUROC                & SAE vs ICA                    & \multicolumn{3}{c}{$p = 0.50$ (tied, $0.715$ vs $0.713$)} \\
AUROC$_\text{scored}$ & SAE vs K-Means / ICA / PCA   & \multicolumn{3}{c}{$\mathbf{p \leq 0.002}$ (each)} \\
Coverage             & SAE vs K-Means                & \multicolumn{3}{c}{$\mathbf{p = 0.007}$} \\
Coverage             & SAE vs ICA\textsuperscript{$\ddagger$}                    & \multicolumn{3}{c}{$p = 0.97$ (ICA wins)} \\
\midrule
\multicolumn{5}{l}{\textit{MovieLens NL audience retrieval (n=32 queries)}} \\
 & & SASRec & BERT4Rec & Stouffer \\
\cmidrule(lr){3-5}
Lift@$n_\text{pos}$  & SAE vs K-Means encoding       & 0.36                  & 0.69                  & 0.54 \\
Lift@$n_\text{pos}$  & SAE vs PCA                    & \textbf{0.0003}       & \textbf{0.0008}       & $\boldsymbol{<10^{-4}}$ \\
Lift@$n_\text{pos}$  & SAE vs ICA                    & \textbf{0.0012}       & \textbf{0.0014}       & $\boldsymbol{<10^{-4}}$ \\
Lift@$n_\text{pos}$  & SAE vs text baseline          & \textbf{0.0070}       & \textbf{0.0028}       & \textbf{0.0001} \\
AUROC                & SAE vs K-Means encoding       & 0.100                 & 0.81                  & 0.41 \\
AUROC                & SAE vs PCA                    & $\boldsymbol{<10^{-4}}$   & $\boldsymbol{<10^{-4}}$   & $\boldsymbol{<10^{-4}}$ \\
AUROC                & SAE vs ICA                    & $\boldsymbol{<10^{-4}}$   & $\boldsymbol{<10^{-4}}$   & $\boldsymbol{<10^{-4}}$ \\
AUROC                & SAE vs text baseline          & 0.895                 & 0.661                 & 0.881 \\
AUROC$_\text{scored}$ & SAE vs K-Means encoding      & \textbf{0.014}        & 0.07                  & \textbf{0.005} \\
AUROC$_\text{scored}$ & SAE vs PCA                   & $\boldsymbol{<10^{-4}}$   & \textbf{0.0001}       & $\boldsymbol{<10^{-4}}$ \\
AUROC$_\text{scored}$ & SAE vs ICA                   & $\boldsymbol{<10^{-4}}$   & $\boldsymbol{<10^{-4}}$   & $\boldsymbol{<10^{-4}}$ \\
AUROC$_\text{scored}$ & SAE vs text baseline         & 0.943                 & 0.688                 & 0.928 \\
Coverage             & SAE vs K-Means encoding       & $\boldsymbol{<10^{-4}}$   & $\boldsymbol{<10^{-4}}$   & $\boldsymbol{<10^{-4}}$ \\
Coverage             & SAE vs PCA                    & \textbf{0.0002}       & \textbf{0.0036}       & $\boldsymbol{<10^{-4}}$ \\
Coverage             & SAE vs ICA                    & \textbf{0.017}        & $\boldsymbol{<10^{-4}}$   & $\boldsymbol{<10^{-4}}$ \\
Coverage             & SAE vs text baseline          & 1.000                 & 1.000                 & 1.000 \\
\midrule
\multicolumn{5}{l}{\textit{Audience characterization (n=20 franchises, Stouffer across encoders)}} \\
Slate redundancy (K-Means$>$SAE)\textsuperscript{$\dagger$}   & median pairwise cosine    & \textbf{0.002} & \textbf{0.049} & $\boldsymbol{\approx 4{\times}10^{-4}}$ \\
Reverse-selector coverage           & SAE vs K-Means encoding   & \textbf{0.0003} & \textbf{0.0001} & $\boldsymbol{\approx 10^{-7}}$ \\
\bottomrule
\end{tabular}
\end{table}

\FloatBarrier
\section{Auto-interpretation and retrieval prompts}\label{app:prompts}

This section lists the LLM prompt templates used at each stage of the pipeline. We use Claude Sonnet~4.6 for feature interpretation (the offline, one-time auto-interp stage) and Claude Haiku~4.5 for query-time feature scoring. The key prompt elements of each stage are summarized below.

\subsection{Feature interpretation (auto-interp)}

For each SAE / baseline latent, we sample 10 top-activating users (positives) and 10 random non-activating users (negatives). For Open e-commerce we show per-window purchase-category summaries. For MovieLens we show a structured per-user behavioral summary (separately cached; see Appendix~\ref{app:movielens-summary-prompt} for the summary prompt used). The interpretation prompt varies by dataset to match what the underlying embedding can encode:

\paragraph{Open e-commerce (SVD-on-counts embeddings based on categories).} The prompt instructs the model that the embedding is a TF-IDF-weighted SVD of user-by-category purchase counts in 6-month windows, and therefore encodes only which categories a user bought and roughly how often. It forbids brand-, title-, price-, or cadence-level claims, since the SAE cannot see them. The model is asked to infer \emph{who these people are} (life-stage / lifestyle / household persona) from category combinations, at a granularity matched to the feature's sparsity (common features $\to$ broader personas; rare features $\to$ specific personas). Output is a JSON object with \texttt{feature\_name}, \texttt{feature\_description}, \texttt{confidence\_score}, and per-pattern reasoning. An enrichment context precomputed over the top-activating users' windows (top categories with lift $>2\times$ vs.\ population) is supplied alongside the per-window examples to anchor the LLM on population-distinctive categories.

\paragraph{MovieLens (transformer embeddings based on titles).} The prompt instructs the model that each positive is a user's behavioral summary computed from their viewing history (see Appendix~\ref{app:movielens-summary-prompt}). The model is asked to identify the specific taste pattern that distinguishes positives from negatives, drawing on title semantics, genre mix, era, and co-viewing patterns that the sequence encoder can encode. Output is the same JSON schema.

Key design choices: (i)~the system prompt forbids hallucinated attributes for the embedding modality, (ii)~positives and negatives are presented with matched formatting so the model cannot detect class via superficial cues, (iii)~confidence scores below $0.6$ are excluded from downstream retrieval.

\subsection{MovieLens behavioral-summary prompt}\label{app:movielens-summary-prompt}

Before auto-interpretation, each MovieLens user is summarized via a one-shot LLM call (Claude Sonnet~4.6) on their chronological watch history. The summary is cached per user and reused across all SAE / baseline auto-interp passes.

\paragraph{System prompt.} We provide the following system prompt: ``You are a careful analyst producing structured behavioral summaries from movie watching histories. Base all claims only on the data shown. Do not invent patterns the history does not support.''

\paragraph{Per-user prompt.} For each user, the prompt supplies the chronologically ordered watch history (up to 150 items; title + release year + genres), basic stats (total count, earliest/latest release year), and instructs the model to produce six terse sections: (i)~\emph{Volume and engagement} (HEAVY/MODERATE/CASUAL classification against a 150-film MovieLens median), (ii)~\emph{Era profile} (decade-level release-year distribution with approximate fractions), (iii)~\emph{Genre profile} (top 5--7 genres with fractions), (iv)~\emph{Stylistic/thematic tendencies} (2--5 specific patterns, avoiding vague labels), (v)~\emph{Director/franchise affinity} (directors with ${\geq}3$ films or clear franchise patterns), (vi)~\emph{Avoidances} (genres/eras notably absent). The prompt explicitly forbids the model from inferring time-viewing rates or absolute viewing dates, since MovieLens does not include viewing timestamps and release years only bracket the films themselves. These summaries comprise the positive and negative examples shown to the feature-interpretation prompt above, which produces each feature's name, description, and confidence score.

\subsection{Query-time feature scoring}

Given an NL audience query, the retrieval pipeline asks the LLM to score each interpreted feature's relevance to the query on a 0--10 scale. The prompt shows the query and a batch of \texttt{\{feature\_name, feature\_description\}} pairs; the model returns a score per feature and optionally a brief rationale. Features scoring $\geq 5$ are retained; the retained features' activations are weighted by a sharpened cosine similarity between the feature text and the query (using \texttt{all-MiniLM-L6-v2} sentence embeddings) combined with the LLM score, capped at 50 features per query, and summed per user. This is the ``weighted sum of activations on retained features'' described in Section~\ref{sec:realdata-setup}.

The weighted-sum combinator is itself compositional: each audience query is answered by combining contributions from multiple LLM-selected features rather than any single one, which is what allows the pipeline to cover both broad audiences (e.g., genre-level) and niche audiences (e.g., era $\times$ sub-genre) without per-query structural changes.

\FloatBarrier

\end{document}